\documentclass[a4paper,fleqn]{cas-sc}

\usepackage[numbers]{natbib}

\usepackage{amssymb}
\usepackage{amsmath}
\usepackage{url}
\usepackage{booktabs}
\usepackage{tabularx}
\usepackage{array}
\usepackage{placeins}   
\usepackage{makecell}
\usepackage{enumitem}
\usepackage{subcaption}

\begin{document}
\shorttitle{ClusLoad: Cluster-Guided Model Sharing}
\shortauthors{Fangzhou Lin et al.}

\title[mode = title]{ClusLoad: Cluster-Guided Model Sharing for Scalable Distribution Transformer Load Forecasting}

\author[1]{Fangzhou Lin}
\fnmark[1]

\author[2]{Guoshun He}
\fnmark[1]

\author[1]{Yi Wang}[orcid=0000-0002-1280-5418]
\cormark[1]
\ead{yiwang123@whu.edu.cn}

\author[3]{S. Supraja}

\fntext[1]{These authors contributed equally to this work.}
\cortext[1]{Corresponding author.}

\affiliation[1]{
    organization={School of Electrical Engineering, Wuhan University},
    addressline={No. 299 Bayi Road, Wuchang District},
    city={Wuhan},
    postcode={430072},
    state={Hubei},
    country={China}
}

\affiliation[2]{
    organization={Institute of Industrial Science, The University of Tokyo},
    addressline={4-6-1 Komaba, Meguro-ku},
    city={Tokyo},
    postcode={153-8505},
    state={Tokyo},
    country={Japan}
}

\affiliation[3]{
    organization={School of Electrical and Electronic Engineering, Nanyang Technological University},
    addressline={50 Nanyang Avenue},
    city={Singapore},
    postcode={639798},
    country={Singapore}
}
\begin{abstract}
Large-scale device-level load forecasting for distribution transformers is
challenged by substantial cross-device heterogeneity and the deployment burden
of maintaining device-specific forecasting models.
Existing approaches typically rely either on device-specific models, which
are difficult to scale to large transformer populations, or on global models,
which reduce deployment complexity but may overlook heterogeneous load
patterns across devices. To address this trade-off, this study develops a
trend-clustered forecasting framework that enables model sharing among
transformers with similar temporal behaviors.
Representation Learning for Time Series using Time-Embeddings (T-Rep)
is employed to learn compact temporal representations of transformer load
series, which are subsequently grouped using a Gaussian mixture model (GMM).
Representative transformers are sampled within each cluster to train
cluster-specific Patch Time Series Transformer (PatchTST) models while
retaining device-level predictions.
The framework is evaluated using real-world 15-min operational data from
nearly one thousand distribution transformers.
The proposed framework reduces mean absolute error (MAE) by approximately
20\% relative to the direct PatchTST baseline, achieves a high-load warning
accuracy of 95.66\%, while requiring only seven cluster-specific forecasting models.
These results demonstrate that behavior-based device organization provides an
effective balance between forecasting accuracy, deployment scalability, and
risk-oriented operational support for distribution transformer management.
\end{abstract}



\maketitle


\section{Introduction}

\subsection{Background}
\label{subsec:background}

With the increasing digitalization and automation of power systems, load forecasting has become an important tool for system operation, equipment management, and risk assessment. Accurate forecasts help utilities anticipate changes in electricity demand and support decisions related to resource allocation, network operation, and system security. Meanwhile, the growing penetration of renewable energy, electric vehicles, and increasingly diverse end-use behaviors has made electricity demand more variable and difficult to predict, placing higher requirements on forecasting accuracy and adaptability.

Among different forecasting levels, distribution transformer load forecasting is particularly relevant to refined distribution-system operation. Distribution transformers form the interface between distribution networks and end users, and their operating conditions directly affect local supply reliability. Fine-grained forecasting at the distribution level can help operators anticipate load peaks and potential network constraints \cite{pinheiro2023secondary}. Prolonged high-load or overload conditions may also lead to excessive temperature rise, accelerated insulation aging, and increased risks of equipment failure and supply interruption \cite{Visakh2023,IEEE_C5791_2025}. Forecasting the future load of individual transformers can therefore provide useful information for preventive measures such as equipment inspection, load transfer, and operational adjustment.

Compared with system- or regional-level forecasting, transformer-level forecasting provides a more detailed view of local operating conditions but also introduces greater modeling complexity. Aggregated demand tends to smooth part of the random variation among individual consumers, whereas transformer loads are more directly affected by local consumption behavior, operating conditions, and network adjustments. More importantly, a practical forecasting system may need to serve hundreds or thousands of transformers simultaneously, whose load patterns can differ substantially from one another. Transformer-level forecasting is therefore not only a time-series prediction problem, but also a problem of how to organize forecasting models across a large and heterogeneous device population.

\subsection{Scalability Challenge}
\label{subsec:scalability_challenge}

In large-scale transformer forecasting, improving the accuracy of an individual model is only part of the problem. As the number of devices increases, the forecasting system must also balance device-level adaptation, model quantity, training cost, and deployment complexity. Existing forecasting strategies generally lie between two extremes: local modeling and global modeling.

Local modeling trains a separate model for each transformer and can therefore preserve device-specific load characteristics. However, the number of models grows with the transformer population, increasing the burden of training, storage, maintenance, and updating. Global modeling instead shares one model across many devices, offering greater parameter and deployment efficiency, but substantial cross-device heterogeneity may make it difficult for a single model to represent all load patterns adequately. Similar trade-offs between local specialization and global model sharing have been reported in large-scale forecasting studies \cite{MonteroManso2021,Godahewa2021}. In distribution-network applications, localization mechanisms have also been explored to improve the ability of global models to accommodate heterogeneous loads \cite{Grabner2023}.

Simple intermediate strategies do not fully resolve this issue. Randomly selecting a subset of transformers can reduce the amount of training data, but the selected devices may not adequately represent the diversity of load behaviors in the full population. Geographical or administrative grouping is convenient for system management, yet spatial proximity does not necessarily imply similar electricity consumption patterns. Transformers within the same region may exhibit substantially different load profiles, while transformers in different regions may show similar temporal behaviors.

These observations suggest that the key question is not simply whether forecasting models should be local or global, but which transformers should share a model. An intermediate modeling strategy should exploit behavioral similarities among devices while retaining sufficient specialization for heterogeneous load patterns. This motivates a data-driven organization of transformers according to their historical load behavior, so that model sharing is determined by temporal similarity rather than by device identity, random selection, or administrative boundaries.

\subsection{Research Gaps and Contributions}
\label{subsec:research_gaps_contributions}

Existing studies provide several important building blocks for large-scale
device-level load forecasting. Fine-grained forecasting has progressed from
system-level demand toward secondary substations and individual consumers
\cite{pinheiro2023secondary,li2021apartment}; load-profile clustering has been
used to identify characteristic electricity-consumption patterns
\cite{motlagh2019clustering,michalakopoulos2024clustering}; and self-supervised
methods such as TS2Vec and T-Rep can learn compact representations of temporal
behavior \cite{yue2022ts2vec,fraikin2024trep}. Transfer, personalized, and
representation-based learning strategies have further demonstrated the value
of sharing information across heterogeneous energy users or buildings
\cite{fan2024pfl,choi2026pptl,ramos2026timevae}. Despite these advances, three
research gaps remain in large-scale distribution transformer-level forecasting:

\begin{enumerate}[label=\arabic*)]
    \item Existing fine-grained forecasting studies are mainly conducted at
    secondary-substation, feeder, building, or individual-consumer levels
    \cite{pinheiro2023secondary,huyghues2020cleansing,li2021apartment}, while
    systematic forecasting across large populations of individual distribution
    transformers remains comparatively less explored. The restricted
    accessibility of utility-operated transformer data further limits the
    evaluation of forecasting methods in realistic large-scale device-level
    settings.

    \item Distribution transformers exhibit substantial differences in their
    temporal load behaviors, yet the appropriate level of model sharing for such
    heterogeneous device populations remains insufficiently explored. Existing
    clustering and representation-based approaches have been used for load
    segmentation, source selection, or representative-client selection
    \cite{motlagh2019clustering,michalakopoulos2024clustering,choi2026pptl,ramos2026timevae},
    but their use as a model-organization mechanism for determining which
    transformers should share forecasting models remains limited. Device-specific
    models are difficult to maintain at scale, whereas a single global model may
    overlook important device-level differences.

    \item Representative load forecasting studies predominantly evaluate
    performance using numerical error metrics
    \cite{pinheiro2023secondary,chitalia2020rnn,gao2023adaptive}, with less
    attention paid to whether forecasts can identify operationally important
    high-load conditions. This leaves a gap between improvements in numerical
    forecasting accuracy and their practical value for distribution-system
    monitoring and risk management.
\end{enumerate}

To address these gaps, this study develops ClusLoad, a cluster-guided
model-sharing framework for large-scale transformer-level load forecasting.
The main contributions are summarized as follows:

\begin{enumerate}[label=\arabic*)]
    \item A large-scale empirical study is conducted using real-world 15-min
    operational data from nearly one thousand distribution transformers across
    23 regions. The study focuses on fine-grained device-level forecasting and
    provides a practical setting for examining cross-device heterogeneity and
    forecasting scalability.

    \item ClusLoad organizes transformers according to their learned temporal
    load behaviors and determines which devices should share forecasting models.
    T-Rep is used to encode historical load characteristics, GMM is employed to
    identify behaviorally similar transformer groups, and cluster-specific
    PatchTST models are trained using representative devices within each cluster.
    This design introduces an intermediate modeling granularity between fully
    local and fully global forecasting, reducing the number of maintained models
    while preserving device-level predictions.

    \item A rated-capacity-based high-load warning evaluation is incorporated
    alongside conventional forecasting metrics. By examining whether predicted
    loads correctly identify high-load operating states, the evaluation links
    numerical forecasting performance with operationally relevant risk
    identification and provides a more application-oriented assessment of
    transformer load forecasting.
\end{enumerate}

\section{Related Work}
\label{sec:related_work}

\subsection{Load Forecasting Across Different Aggregation Levels}
\label{subsec:aggregation_levels}

\subsubsection{System-Level and Regional-Level Load Forecasting}
Load forecasting has been extensively studied at system and regional levels because it provides essential information for generation scheduling, market operation, network planning, and security assessment. At these aggregation levels, electricity demand represents the combined behavior of a large number of consumers. Consequently, part of the stochastic variation of individual users is averaged out, and the resulting load series generally exhibit clearer periodicity and smoother temporal structures. Haben et al. reviewed low-voltage load forecasting methods and emphasized that forecasting requirements, data properties, and useful modeling strategies change substantially with the spatial and electrical aggregation level \cite{haben2021review}. Li et al. further demonstrated this heterogeneity by constructing more than 800 distribution-system forecasting tasks with different temporal and aggregation characteristics, showing that no single forecasting model is uniformly preferable across all tasks \cite{li2021meta}.

The distinction between aggregation levels is also important from an operational perspective. System-level forecasts are primarily used for high-level scheduling and balancing, whereas lower-level forecasts can support more localized network decisions. Pinheiro et al. developed a forecasting framework spanning the national system and approximately 100,000 secondary substations in the Portuguese distribution network, demonstrating both the feasibility and the practical value of extending forecasting toward finer network levels \cite{pinheiro2023secondary}. Their work illustrates an important transition in modern power-system forecasting: as monitoring becomes more granular, the forecasting problem increasingly shifts from estimating an aggregated demand trajectory to simultaneously characterizing a large population of local operating units.

\subsubsection{Low-Voltage and Secondary-Substation Load Forecasting}
Forecasting at the distribution level introduces data and modeling challenges that are less pronounced in highly aggregated load series. Huyghues-Beaufond et al. studied short-term forecasting for 342 medium-voltage feeders and showed that real-world distribution data may contain gross errors, missing observations, and structural breaks caused by network reconfiguration and load transfers \cite{huyghues2020cleansing}. These phenomena are directly relevant to practical forecasting systems because a model trained on historical operating conditions may become less representative when network topology or local load allocation changes.

The growing availability of smart-meter and substation measurements has consequently motivated research on low-voltage forecasting at increasingly fine granularity. However, the review by Haben et al. noted that the low-voltage domain remains substantially less standardized than traditional system-level forecasting, particularly with respect to datasets, applications, and reproducible evaluation \cite{haben2021review}. Pinheiro et al. similarly emphasized applicability and reproducibility when extending short-term forecasting from the system level to secondary substations \cite{pinheiro2023secondary}. These studies provide an important basis for large-scale transformer forecasting, but secondary-substation or feeder forecasting still aggregates the behavior of multiple downstream users and does not fully characterize the heterogeneity that appears when individual distribution devices are treated as independent forecasting targets.

\subsubsection{Fine-Grained Device-Level Forecasting}
As the forecasting target becomes more localized, load series generally become more volatile and more strongly affected by individual consumption behaviors. Li et al. reported that apartment-level electricity demand contains pronounced volatility and idiosyncratic variations and compared forecasting performance across apartment, floor, and building aggregation levels \cite{li2021apartment}. Such observations support the general view that aggregation can partially smooth irregular individual behavior, whereas fine-grained forecasting requires models to cope with stronger local variability.

Transformer-level forecasting represents a particularly relevant device-level task in distribution networks because the transformer is both a measurable asset and an operationally meaningful unit. Compared with household forecasting, transformer forecasting reflects the aggregated behavior of a local customer group while remaining sufficiently fine-grained to support equipment-oriented monitoring, load transfer, and maintenance decisions. Nevertheless, forecasting a single transformer is only part of the practical problem. A utility may operate hundreds or thousands of transformers simultaneously, so a useful forecasting framework must also address cross-device heterogeneity, model management, and deployment cost. This large-population perspective motivates the scalable modeling strategies discussed in Section~\ref{subsec:scalable_forecasting}.

\subsection{Deep Learning for Load and Time-Series Forecasting}
\label{subsec:deep_forecasting}

\subsubsection{Deep Learning-Based Load Forecasting}
Deep neural networks have become widely used in load forecasting because of their ability to learn nonlinear temporal dependencies directly from historical measurements. Chitalia et al. investigated recurrent neural-network and clustering hybrids for commercial-building load forecasting and reported that deep recurrent models can provide robust forecasting performance across different building types, locations, and sampling resolutions \cite{chitalia2020rnn}. Their study also showed that clustering can improve forecasting when different operating patterns are present, suggesting that identifying behavioral structure before prediction can be beneficial.

More recent research has focused on adaptability across locations and operating conditions. Gao et al. combined Transformer-based forecasting with domain knowledge, transfer learning across locations, and online learning across time, thereby addressing changes in data distributions and limited local data \cite{gao2023adaptive}. These developments demonstrate that forecasting accuracy alone is no longer the only objective. For realistic energy applications, models are increasingly expected to generalize across heterogeneous sites, adapt to distribution shifts, and remain computationally manageable when deployed at scale.

\subsubsection{Transformer-Based Time-Series Forecasting}
Transformer architectures have significantly expanded the design space of long-horizon time-series forecasting. Informer introduced ProbSparse self-attention and an efficient generative decoder to reduce the computational burden of long-sequence forecasting \cite{zhou2021informer}. Pyraformer constructed pyramidal multi-resolution attention to model temporal dependencies at different scales with linear complexity in sequence length \cite{liu2022pyraformer}. FEDformer combined seasonal--trend decomposition with frequency-domain representations to capture global temporal structures efficiently \cite{zhou2022fedformer}, while FiLM used Legendre projections and frequency-domain filtering to preserve long historical information while suppressing noise \cite{zhou2022film}. TimesNet reformulated temporal variation modeling in a two-dimensional representation space so that multiple periodicities could be captured by a unified backbone \cite{wu2023timesnet}.

At the same time, the rapid development of complex forecasting architectures has motivated reassessment of model complexity. Zeng et al. showed that simple linear models can remain highly competitive in long-term forecasting and proposed the DLinear family as a strong lightweight baseline \cite{zeng2023linear}. More recently, iTransformer inverted the conventional Transformer tokenization strategy and represented individual variates as tokens, allowing self-attention to model cross-variate correlations while feed-forward networks learn representations within each variate \cite{liu2024itransformer}. These models provide diverse and representative baselines for evaluating whether improvements arise from the forecasting backbone itself or from the proposed organization of heterogeneous transformer devices.

\subsubsection{Patch-Based and Channel-Independent Forecasting}
PatchTST is particularly relevant to the present study because it combines patch-based tokenization with channel-independent forecasting \cite{nie2023patchtst}. Instead of treating every time point as an individual token, PatchTST divides a time series into local patches. This reduces the effective token length, lowers attention cost, and allows the model to exploit a longer historical context. Its channel-independent design further processes each channel as a univariate sequence while sharing embedding and Transformer parameters across channels, which provides an efficient mechanism for learning from multiple series.

These characteristics make PatchTST a suitable forecasting backbone for a population of transformer load series. However, parameter sharing at the channel level does not itself determine which heterogeneous devices should be trained together. If transformers with substantially different load dynamics are indiscriminately combined, a shared model may still be required to fit conflicting patterns. The present work therefore treats PatchTST as the supervised forecasting backbone and addresses a complementary problem before forecasting: how to organize a large population of transformers according to their temporal behavior so that parameter sharing occurs among devices with more compatible load patterns.

\subsection{Time-Series Representation Learning and Load-Pattern Clustering}
\label{subsec:representation_clustering}

\subsubsection{Conventional Load-Profile Clustering}
Load-profile clustering has long been used to identify groups of consumers with similar electricity-use patterns. Rhodes et al. applied K-means clustering to residential electricity demand profiles and derived representative seasonal usage groups \cite{rhodes2014clustering}. Motlagh et al. highlighted a central difficulty of directly clustering long electricity time series: the extreme dimensionality, noise, and unequal quality of raw load observations can make similarity estimation unreliable. They therefore transformed long load series into compact model-based representations before clustering customers \cite{motlagh2019clustering}. More recently, Michalakopoulos et al. systematically compared K-means, K-medoids, hierarchical agglomerative clustering, and DBSCAN for residential load-profile segmentation and emphasized the need to examine both cluster number and within-cluster similarity \cite{michalakopoulos2024clustering}.

Clustering has also been incorporated directly into forecasting systems. Li et al. proposed a clustering-based cross-scale load prediction framework in which representative cluster information was used to construct building-level predictions from thermal-zone loads \cite{li2021crossscale}. Dong et al. used K-medoids with derivative dynamic time warping to identify electricity-consumption-behavior similarity across building partitions and floors before applying spatiotemporal forecasting models \cite{dong2025spatiotemporal}. These studies demonstrate that behavior-based grouping can improve the organization of energy forecasting tasks. Nevertheless, conventional load-profile clustering often relies on raw curves, handcrafted features, fixed daily profiles, or task-specific distance measures, which can be sensitive to scale, noise, and the selected temporal window.

\subsubsection{Self-Supervised Time-Series Representation Learning}
Self-supervised representation learning provides an alternative way to characterize long and noisy time series. Instead of defining similarity directly in the raw observation space, an encoder maps each series to a compact latent representation that preserves informative temporal structure. TS2Vec introduced hierarchical contrastive learning to obtain contextual representations at different temporal resolutions and demonstrated that the learned representations can support forecasting, classification, and anomaly detection \cite{yue2022ts2vec}. Such approaches are attractive for large transformer populations because they allow device similarity to be evaluated in a learned feature space rather than by point-wise comparison of long load curves.

T-Rep further incorporates explicit time embeddings into self-supervised representation learning \cite{fraikin2024trep}. Its design is intended to capture temporal features such as trends, periodicities, and distribution shifts while improving robustness to noisy or incomplete observations. These properties are particularly relevant to transformer load data, where the objective of clustering is not to match every short-term fluctuation but to identify devices with similar underlying operating patterns. Accordingly, the proposed framework employs T-Rep to encode each transformer history before clustering, thereby separating the representation-learning problem from the downstream forecasting task.

\subsubsection{Clustering in Learned Representation Spaces}
Recent energy-forecasting studies increasingly use learned latent representations to measure similarity across sites. Choi et al. proposed a privacy-preserving transfer-learning framework in which an unsupervised contrastive encoder maps anonymized building time series into latent vectors, and representation similarity is then used to select source buildings for transfer learning \cite{choi2026pptl}. This work is important because it demonstrates that temporal dynamics alone can provide a meaningful basis for cross-site similarity even when metadata such as building type or location are unavailable.

A closely related direction is representative selection from latent spaces. Ramos et al. used TimeVAE latent representations to cluster electricity-demand clients and select representative participants in a federated forecasting framework, reducing participation cost relative to full-client training and avoiding the pattern-coverage uncertainty of random selection \cite{ramos2026timevae}. These studies provide strong evidence for the usefulness of representation-based similarity. However, their objectives differ from the present work: Choi et al. focus on source selection for transfer learning, while Ramos et al. focus on client participation in federated learning. The present study instead uses T-Rep embeddings and GMM clustering to organize a centralized population of distribution transformers into forecasting groups, retain cluster prototypes for later device assignment, and train cluster-level forecasting models for all devices in each group.

\subsection{Scalable Forecasting for Heterogeneous Device Populations}
\label{subsec:scalable_forecasting}

\subsubsection{Local and Global Forecasting Strategies}
A large population of forecasting targets naturally creates a choice between local and global modeling. Local models are trained independently for individual devices and can closely adapt to device-specific characteristics, but their computational, storage, updating, and maintenance costs increase approximately with the number of devices. This limitation becomes important when the forecasting system contains hundreds or thousands of transformers and models must be periodically updated.

Global modeling takes the opposite approach by sharing one model across many series. It reduces model quantity and can exploit a larger combined dataset, but its effectiveness depends on whether the participating series share sufficiently compatible patterns. Distribution-system forecasting tasks are heterogeneous across aggregation levels and operating conditions \cite{li2021meta}, and fine-grained electricity series can exhibit strong local idiosyncrasies \cite{li2021apartment}. Therefore, a single global model may underrepresent minority or atypical patterns even when its average forecasting accuracy is satisfactory. The practical problem is consequently not a binary choice between local and global models, but how to determine an intermediate level of model sharing that preserves important device differences without reproducing the management burden of fully local modeling.

\subsubsection{Transfer and Personalized Learning Across Devices}
Transfer learning and federated learning provide alternative mechanisms for sharing information across related energy forecasting tasks. Fan et al. showed that knowledge learned from well-measured buildings can improve short-term predictions for buildings with limited local data, demonstrating the value of cross-site information transfer \cite{fan2020transfer}. However, transfer effectiveness depends strongly on the relationship between source and target domains, and inappropriate source selection can lead to negative transfer.

The same heterogeneity issue appears in federated learning. Fan et al. investigated personalized federated learning for cross-building energy prediction and showed that differences in building operation and energy patterns can limit standard global federated models, while personalization can recover performance across heterogeneous participants \cite{fan2024pfl}. These findings reinforce the central scalability dilemma considered in this study: shared learning is efficient, but unrestricted sharing may be inappropriate when devices exhibit substantially different temporal behaviors. Rather than personalizing a separate model for every transformer, the proposed framework seeks a coarser but data-driven level of personalization by assigning transformers with similar representations to the same forecasting cluster.

\subsubsection{Cluster-Based Representative Modeling}
Cluster-based modeling offers a natural intermediate strategy between fully local and fully global forecasting. Li et al. demonstrated that representative cluster information can support cross-scale building-load prediction and explicitly observed a trade-off between cluster sample size and prediction accuracy \cite{li2021crossscale}. Chitalia et al. also reported benefits from combining clustering with deep recurrent forecasting for commercial buildings \cite{chitalia2020rnn}. These studies indicate that grouping related load patterns can reduce unnecessary heterogeneity within a forecasting task.

Representation-based representative selection has recently extended this idea to larger distributed populations. Ramos et al. showed that latent-space clustering can identify representative electricity-demand clients more effectively than random participation in a federated setting \cite{ramos2026timevae}. In the present transformer-level setting, representative sampling serves a different purpose: a limited number of transformers from each learned load-pattern cluster are used to train cluster-specific PatchTST models, and the resulting model is then applied to all transformers assigned to that cluster. This design reduces the number and scale of forecasting models while retaining a data-driven distinction between heterogeneous device groups. Moreover, by retaining GMM cluster prototypes, the framework can assign newly introduced or reconfigured transformers to an existing forecasting model without immediately constructing a new device-specific model.

\begin{table}[pos=htbp]
\centering
\caption{Summary of representative methods related to large-scale load forecasting.}
\label{tab:related_work_comparison}

\footnotesize
\setlength{\tabcolsep}{4pt}
\renewcommand{\arraystretch}{1.12}

\begin{tabularx}{\textwidth}{
    >{\raggedright\arraybackslash}p{2.35cm}
    >{\raggedright\arraybackslash}p{2.85cm}
    >{\raggedright\arraybackslash}X
    >{\raggedright\arraybackslash}X}
\toprule
\textbf{Research direction} &
\textbf{Representative studies} &
\textbf{Core idea} &
\textbf{Main limitation} \\
\midrule

Deep forecasting models &
Informer \cite{zhou2021informer};
FEDformer \cite{zhou2022fedformer};
TimesNet \cite{wu2023timesnet};
iTransformer \cite{liu2024itransformer};
PatchTST \cite{nie2023patchtst} &
Model temporal dependencies using attention, frequency-domain methods, or patch-based representations. &
Focus mainly on forecasting architectures rather than organizing heterogeneous devices. \\

\addlinespace

Load-profile clustering &
Rhodes et al. \cite{rhodes2014clustering};
Motlagh et al. \cite{motlagh2019clustering};
Michalakopoulos et al. \cite{michalakopoulos2024clustering} &
Group loads according to similarities in their consumption patterns. &
Clustering raw profiles or handcrafted features can be sensitive to noise, scale, and dimensionality. \\

\addlinespace

Time-series representation learning &
TS2Vec \cite{yue2022ts2vec};
T-Rep \cite{fraikin2024trep} &
Map long time series into compact latent representations of temporal behavior. &
Does not by itself determine how forecasting models should be shared across devices. \\

\addlinespace

Transfer and personalized learning &
Fan et al. \cite{fan2020transfer};
Fan et al. \cite{fan2024pfl};
Choi et al. \cite{choi2026pptl} &
Share or transfer knowledge across related forecasting targets while accounting for heterogeneity. &
Often requires source selection, personalization, or device-specific adaptation. \\

\addlinespace

Cluster-based representative forecasting &
Li et al. \cite{li2021crossscale};
Dong et al. \cite{dong2025spatiotemporal};
Ramos et al. \cite{ramos2026timevae} &
Group similar loads and use representative devices or cluster-level information for forecasting. &
Few studies jointly address learned clustering, scalable transformer-level forecasting, and model sharing. \\

\bottomrule
\end{tabularx}
\end{table}
\FloatBarrier

\subsection{Research Positioning}
\label{subsec:research_positioning}

Against this background, the proposed framework is positioned as a scalable
organization-and-forecasting strategy rather than as a replacement for existing
forecasting backbones. Historical transformer loads are first encoded with
T-Rep to obtain temporal representations, after which GMM is used to identify
groups of devices with similar load dynamics. Representative transformers from
each cluster are used to train cluster-level PatchTST forecasting models. This
creates an intermediate modeling granularity between one-model-per-device and
one-global-model-for-all, with the objective of balancing device heterogeneity
against training and deployment efficiency.

The framework further retains cluster prototypes so that newly introduced or
reconfigured transformers can be mapped to an existing forecasting group
according to their recent temporal representation. In addition to conventional
MAE-, MSE-, RMSE-, MAPE-, and MSPE-based evaluation, an engineering-oriented
rated-capacity warning task is used to quantify whether the forecasts identify
high-load conditions. The combination of large-scale transformer-level data,
learned temporal organization, cluster-level forecasting, reusable model
assignment, and operational warning evaluation defines the specific research
position of this study relative to the existing literature.

\section{Methodology}
\label{sec:methodology}

\subsection{Problem Formulation}
\label{subsec:problem_formulation}

This section formulates the transformer-level forecasting problem and
introduces the modeling granularities considered in this study. For reference,
the main notation used throughout the proposed framework is summarized in
Table~\ref{tab:notation}, while each variable is also defined when it first
appears in the text.

\begin{table}[pos=htbp]
\centering
\caption{Main notation used throughout the proposed framework.}
\label{tab:notation}
\small
\setlength{\tabcolsep}{5pt}
\renewcommand{\arraystretch}{1.15}

\begin{tabularx}{\textwidth}{
    >{\raggedright\arraybackslash}p{2.0cm}
    >{\raggedright\arraybackslash}X}
\toprule
\textbf{Symbol} & \textbf{Definition} \\
\midrule
$N$ & Number of transformers used for model construction and forecasting \\
$L_r$ & Historical sequence length used for temporal representation learning \\
$L$ & Forecasting input-window length \\
$T$ & Forecasting horizon \\
$F$ & Number of input variables for one transformer \\
$d$ & Dimension of the T-Rep device representation \\
$K$ & Number of GMM clusters \\
$\mathbf{e}_i$ & T-Rep representation of transformer $i$ \\
$\mathbf{z}_i$ & $\ell_2$-normalized representation used for clustering \\
$c_i$ & Cluster label assigned to transformer $i$ \\
$\mathcal{C}_k$ & Set of transformers assigned to cluster $k$ \\
$\boldsymbol{\mu}_k$ & Mean vector (prototype) of GMM component $k$ \\
$\mathcal{S}_k$ & Sampled training-transformer set for cluster $k$ \\
$m_k$ & Number of sampled training transformers in cluster $k$ \\
$f_k$ & PatchTST forecasting model associated with cluster $k$ \\
$C_i$ & Rated capacity of transformer $i$ \\
\bottomrule
\end{tabularx}
\end{table}

\subsubsection{Transformer-Level Forecasting Task}

Consider a population of $N$ distribution transformers indexed by
$i\in\{1,\ldots,N\}$. At forecasting origin $t$, let
$\mathbf{x}_{i,t}\in\mathbb{R}^{F}$ denote the observed feature vector of
transformer $i$, where $F$ is the number of input feature channels. Let
$y_{i,t}\in\mathbb{R}$ denote the observed transformer load at time $t$; this
target load is one component of $\mathbf{x}_{i,t}$, while the remaining
components, when available, are auxiliary variables.

Given an input-window length $L$, the forecasting input
$\mathbf{X}_{i,t}$ contains the most recent $L$ feature vectors up to and
including time $t$:
\begin{equation}
\mathbf{X}_{i,t}
=
[\mathbf{x}_{i,t-L+1},\ldots,\mathbf{x}_{i,t}]
\in\mathbb{R}^{L\times F}.
\end{equation}

The corresponding observed future load trajectory is
$\mathbf{y}_{i,t}=[y_{i,t+1},\ldots,y_{i,t+T}]\in\mathbb{R}^{T}$, where $T$
is the forecasting horizon. The forecasting task is to estimate this trajectory
as
\begin{equation}
\hat{\mathbf{y}}_{i,t}
=
[\hat{y}_{i,t+1},\ldots,\hat{y}_{i,t+T}]
\in\mathbb{R}^{T}.
\end{equation}

The final output therefore preserves device-level resolution: each transformer
receives its own forecast rather than an aggregate regional prediction.

The proposed framework separates two temporal scales. A long historical
sequence is used to characterize the operating pattern of each transformer for
representation learning and clustering, whereas fixed-length sliding windows
are used by the supervised forecasting model. Let $L_r$ denote the length of
the historical sequence used for representation learning. The representation
stage maps each device history to a compact vector, while the forecasting
stage maps local windows $\mathbf{X}_{i,t}$ to future load trajectories.

\subsubsection{Local, Global, and Cluster-Level Modeling}
Three modeling granularities can be distinguished. A local strategy trains an independent forecasting function $f_i(\cdot;\boldsymbol{\theta}_i)$ for every transformer,
\begin{equation}
\hat{\mathbf{y}}_{i,t}=f_i(\mathbf{X}_{i,t};\boldsymbol{\theta}_i),
\qquad i=1,\ldots,N,
\end{equation}
which preserves device specificity but requires $N$ independently maintained models. Here, $\boldsymbol{\theta}_i$ denotes the learnable parameter set of the local model for transformer $i$. At the opposite extreme, a global strategy uses one shared model,
\begin{equation}
\hat{\mathbf{y}}_{i,t}=f_{\mathrm{g}}(\mathbf{X}_{i,t};\boldsymbol{\theta}_{\mathrm{g}}),
\end{equation}
for the full transformer population. The symbol $\boldsymbol{\theta}_{\mathrm{g}}$ denotes the parameter set of the global model. This reduces model quantity but forces heterogeneous operating patterns to share the same parameterization.

The proposed strategy introduces an intermediate, cluster-level granularity. Let the transformer population be divided into $K$ load-pattern clusters $\{\mathcal{C}_1,\ldots,\mathcal{C}_K\}$, and let $c_i\in\{1,\ldots,K\}$ denote the cluster assigned to transformer $i$. One forecasting model is trained for each cluster, yielding
\begin{equation}
\hat{\mathbf{y}}_{i,t}
=f_{c_i}(\mathbf{X}_{i,t};\boldsymbol{\theta}_{c_i}).
\label{eq:cluster_forecast}
\end{equation}
Here, $\boldsymbol{\theta}_{c_i}$ denotes the parameter set of the cluster model selected by the label $c_i$. Accordingly, the model quantity is reduced from $N$ to $K$, while transformers with substantially different temporal behavior are not forced into a single forecasting model.

\subsubsection{Learning Objective}
For a cluster-specific optimization batch containing $B$ forecasting
samples, let $b\in\{1,\ldots,B\}$ index the sample and
$h\in\{1,\ldots,T\}$ index the forecast-horizon step. The supervised
forecasting objective is the mean squared error (MSE):

\begin{equation}
\mathcal{L}_{\mathrm{MSE}}
=
\frac{1}{BT}
\sum_{b=1}^{B}
\sum_{h=1}^{T}
\left(y_{b,h}-\hat{y}_{b,h}\right)^2.
\label{eq:mse_loss}
\end{equation}

The representation model and the forecasting models are trained separately:
T-Rep first learns device-level temporal representations in a self-supervised
manner, after which each cluster-specific PatchTST is optimized using the
above objective.

\subsection{Overall Framework}
\label{subsec:overall_framework}

The proposed framework is designed to organize a large population of
heterogeneous distribution transformers according to their historical load
behaviors before forecasting. Instead of training an independent model for
each transformer or forcing all transformers to share a single global model,
the framework first identifies groups of devices with similar temporal
patterns and subsequently trains forecasting models at the cluster level.

\begin{figure}[htbp]
    \centering
    \includegraphics[
        width=\textwidth
    ]{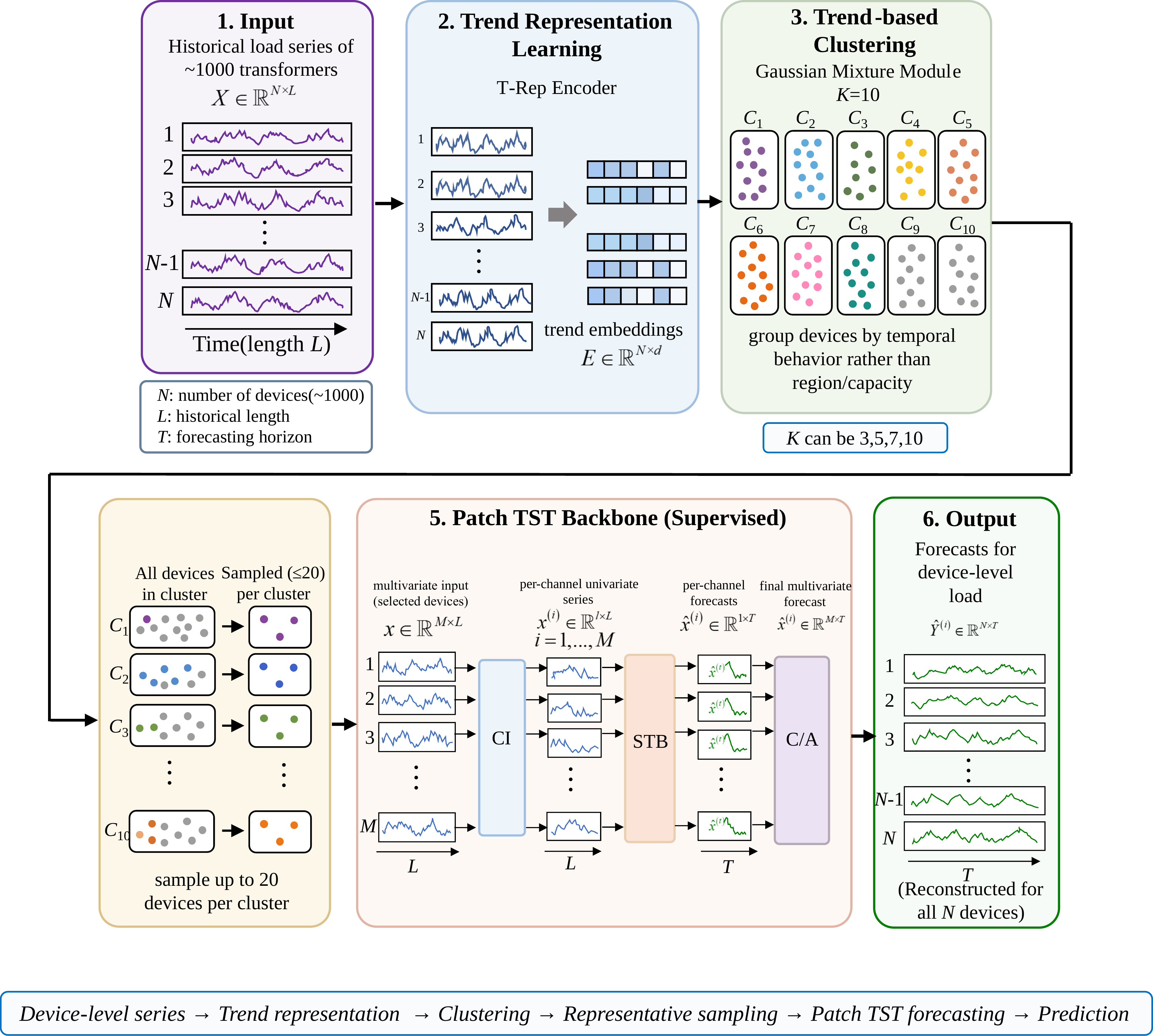}
    \caption{Overall workflow of the proposed transformer-level load
    forecasting framework. Historical load series are encoded by T-Rep and
    grouped using GMM. Representative transformers are selected within each
    cluster to train cluster-specific forecasting models, which are subsequently
    used to generate device-level load forecasts.}
    \label{fig:overall_framework}
\end{figure}

The overall workflow is illustrated in
Fig.~\ref{fig:overall_framework}. The framework links temporal representation
learning, GMM-based clustering, representative sampling, cluster-specific
forecasting, and device-level prediction in a sequential pipeline. Detailed
operations within the PatchTST forecasting model are presented separately in
Section~\ref{subsec:cluster_patchtst}.

\FloatBarrier

The stages in Fig.~\ref{fig:overall_framework} are connected through
explicit intermediate outputs. In Stage~1, the long historical load sequence
$\mathbf{x}^{(r)}_i$ provides the input used to characterize transformer $i$.
Stage~2 maps this sequence through T-Rep to a fixed-dimensional embedding
$\mathbf{e}_i$, which is subsequently normalized to $\mathbf{z}_i$. The
normalized representation becomes the input to the GMM in Stage~3, whose
output is the cluster assignment $c_i$ and the corresponding device group
$\mathcal{C}_{c_i}$. Thus, the first three stages determine which transformers
exhibit similar temporal behavior and should share a forecasting model.

The cluster assignments then determine how the supervised forecasting
models are constructed. In Stage~4, a representative subset
$\mathcal{S}_k\subseteq\mathcal{C}_k$ is selected from each cluster, and the
forecasting windows generated from these transformers form the corresponding
cluster-specific training set $\mathcal{D}_k$. Stage~5 uses
$\mathcal{D}_k$ to train one PatchTST model $f_k$ for cluster $k$. During
Stage~6, transformer $i$ invokes the model associated with its cluster,
$f_{c_i}$, while its own recent input window $\mathbf{X}_{i,t}$ is used to
generate the device-specific forecast
$\hat{\mathbf{y}}_{i,t}=f_{c_i}(\mathbf{X}_{i,t})$. Consequently, clustering
determines \emph{which devices share model parameters}, whereas forecasting
remains at the individual-transformer level.

\subsubsection{Offline Model Construction}

The offline phase constructs the device organization and the associated
cluster-specific forecasting models described in
Fig.~\ref{fig:overall_framework}. The complete mapping is summarized as

\begin{equation}
\begin{aligned}
\{\mathbf{x}^{(r)}_i\}_{i=1}^{N}
&\xrightarrow{\mathrm{T\mbox{-}Rep}}
\{\mathbf{e}_i\}_{i=1}^{N}
\xrightarrow{\ell_2\ \mathrm{norm.}}
\{\mathbf{z}_i\}_{i=1}^{N}
\xrightarrow{\mathrm{GMM}}
\{\mathcal{C}_k\}_{k=1}^{K},
\\
\{\mathcal{C}_k\}_{k=1}^{K}
&\xrightarrow{\mathrm{sampling}}
\{\mathcal{S}_k\}_{k=1}^{K}
\xrightarrow{\mathrm{PatchTST}}
\{f_k(\cdot;\boldsymbol{\theta}_k)\}_{k=1}^{K}.
\end{aligned}
\label{eq:offline_flow}
\end{equation}

Here, $\mathbf{x}^{(r)}_i$ denotes the historical load sequence of
transformer $i$, $\mathbf{e}_i$ its T-Rep embedding, $\mathbf{z}_i$ the
normalized representation used for clustering, $\mathcal{C}_k$ the resulting
cluster, $\mathcal{S}_k$ its representative training subset, and
$f_k(\cdot;\boldsymbol{\theta}_k)$ the PatchTST model trained for that
cluster.

\subsubsection{Online Forecasting and Model Assignment}
For an existing transformer, the offline clustering stage provides its cluster label $c_i$. The corresponding forecasting model $f_{c_i}$ is loaded and applied directly to the transformer's own historical input window. Importantly, a transformer does not need to have participated in the supervised training subset to be forecast: all devices assigned to the same cluster reuse the same cluster-specific PatchTST.

The same organization principle also provides a deployment rule for a transformer that was not included during offline model construction. Its recent history is encoded using the trained T-Rep encoder and processed with the same representation normalization. The resulting representation is matched to the retained cluster prototypes, after which the corresponding cluster model can be used without training a new device-specific forecaster. Section~\ref{subsec:new_device_assignment} formalizes this mechanism.

\subsection{Temporal Representation Learning with T-Rep}
\label{subsec:trep}

\subsubsection{Historical Load Sequence Construction}
For representation learning, each transformer is described by a long
univariate load sequence
\begin{equation}
\mathbf{x}^{(r)}_i=
[x^{(r)}_{i,1},\ldots,x^{(r)}_{i,L_r}]^{\mathsf{T}}
\in\mathbb{R}^{L_r},
\end{equation}
where the superscript $(r)$ distinguishes the representation-learning
sequence from the shorter supervised forecasting window, and $L_r$ denotes
its length. The representation sequence is standardized independently for
each transformer using statistics computed from that transformer's
training-period history. Let $\mu_i^{(r)}$ and $\sigma_i^{(r)}>0$ denote the
corresponding mean and standard deviation. The standardized series is
\begin{equation}
\tilde{x}^{(r)}_{i,t}
=\frac{x^{(r)}_{i,t}-\mu_i^{(r)}}{\sigma_i^{(r)}}.
\label{eq:device_standardization}
\end{equation}
for $t=1,\ldots,L_r$. Device-wise standardization prevents absolute load magnitude from dominating the representation-space geometry and allows the subsequent clustering step to focus more directly on temporal shape and operating-pattern characteristics. The exact observation period and data-screening procedure are specified in Section~4.

\subsubsection{T-Rep Encoding}

T-Rep is an existing self-supervised time-series representation learning
method proposed in \cite{fraikin2024trep}, which combines instance-level and
temporal contrastive learning with time-embedding prediction objectives. In
this study, we adopt T-Rep as the temporal representation component of the
proposed framework rather than introducing a new representation-learning
architecture. Its role is adapted to the transformer-level forecasting setting:
the encoder is trained on the historical load sequences of the transformer
population and is subsequently used to generate one compact device-level
representation for each transformer. These representations are then passed to
the GMM-based clustering stage to organize transformers according to their
temporal load behavior.

Accordingly, T-Rep is trained directly on the transformer representation
dataset rather than using an externally pretrained encoder. Denoting the
T-Rep encoder by $g_{\phi}(\cdot)$, its training objective follows the original
formulation and can be written as
\begin{equation}
\mathcal{L}_{\mathrm{TRep}}
=\lambda_{\mathrm{ins}}\mathcal{L}_{\mathrm{ins}}
+\lambda_{\mathrm{temp}}\mathcal{L}_{\mathrm{temp}}
+\lambda_{\mathrm{jsd}}\mathcal{L}_{\mathrm{jsd}}
+\lambda_{\mathrm{cond}}\mathcal{L}_{\mathrm{cond}},
\label{eq:trep_objective}
\end{equation}
Here, $\mathcal{L}_{\mathrm{ins}}$ and
$\mathcal{L}_{\mathrm{temp}}$ are the instance- and temporal-contrastive
losses inherited from the original T-Rep formulation,
$\mathcal{L}_{\mathrm{jsd}}$ is the time-embedding divergence-prediction loss
based on the Jensen--Shannon divergence (JSD), and
$\mathcal{L}_{\mathrm{cond}}$ is the time-embedding-conditioned forecasting
loss. The coefficients
$\lambda_{\mathrm{ins}},\lambda_{\mathrm{temp}},
\lambda_{\mathrm{jsd}},\lambda_{\mathrm{cond}}\ge 0$ are their scalar loss
weights. Their numerical values and the remaining T-Rep hyperparameters are
reported in Section~4.

After training, one device-level vector is extracted using T-Rep's native
full-series encoding mode. To make this operation explicit, we denote by
$g_{\phi}^{\mathrm{FS}}(\cdot)$ the trained T-Rep encoder operated in
full-series (FS) mode, which returns a single $d$-dimensional representation
for the entire input sequence rather than a representation at each timestamp.
Thus,
\begin{equation}
\mathbf{e}_i
=g_{\phi}^{\mathrm{FS}}(\tilde{\mathbf{x}}^{(r)}_i)
\in\mathbb{R}^{d}.
\label{eq:trep_embedding}
\end{equation}

\subsubsection{Device-Level Trend Representation}

Before clustering, the full-series embedding is normalized to unit $\ell_2$ norm,
\begin{equation}
\mathbf{z}_i
=\frac{\mathbf{e}_i}{\|\mathbf{e}_i\|_2}.
\label{eq:l2_embedding}
\end{equation}

This normalization is applied because the magnitude of the learned
embeddings can vary across transformers and may otherwise influence the
subsequent clustering independently of their latent temporal patterns.
Projecting all embeddings onto the unit hypersphere reduces such
magnitude-induced variation and encourages the GMM to organize transformers
primarily according to the relative structure of their learned representations.
This step complements the device-wise standardization applied to the raw load
series: the former operates in the learned representation space, whereas the
latter operates in the original input space.

The resulting set
$\mathcal{Z}=\{\mathbf{z}_i\}_{i=1}^{N}$ forms the latent device space used
by the clustering stage. The representation is intended to preserve the
temporal characteristics learned by T-Rep while providing a more comparable
basis for clustering across transformers. The clustering algorithm therefore
operates on normalized device-level representations rather than directly on
high-dimensional raw load curves.

\subsection{GMM-Based Trend Clustering}
\label{subsec:gmm_clustering}

\subsubsection{Gaussian Mixture Modeling}
The normalized representations are modeled using a $K$-component Gaussian mixture model. The probability density of a representation $\mathbf{z}_i$ is
\begin{equation}
p(\mathbf{z}_i)
=\sum_{k=1}^{K}\pi_k
\mathcal{N}(\mathbf{z}_i\mid\boldsymbol{\mu}_k,\boldsymbol{\Sigma}_k),
\label{eq:gmm_density}
\end{equation}
Here, $\pi_k\ge0$ is the mixture weight,
$\boldsymbol{\mu}_k\in\mathbb{R}^{d}$ is the component mean, and
$\boldsymbol{\Sigma}_k\in\mathbb{R}^{d\times d}$ is the covariance matrix of
component $k$, with $\sum_{k=1}^{K}\pi_k=1$. The GMM parameters are estimated
by expectation--maximization through maximum-likelihood fitting of
$\mathcal{Z}$. In the implemented configuration,
$\boldsymbol{\Sigma}_k$ is diagonal for every component; the corresponding
regularization, initialization, iteration, and random-state settings are
reported in Section~4.

\subsubsection{Cluster Assignment}
For each transformer, the posterior responsibility of component $k$ is
\begin{equation}
\gamma_{ik}
=\Pr(c_i=k\mid\mathbf{z}_i)
=\frac{\pi_k\mathcal{N}(\mathbf{z}_i\mid\boldsymbol{\mu}_k,\boldsymbol{\Sigma}_k)}
{\sum_{j=1}^{K}\pi_j\mathcal{N}(\mathbf{z}_i\mid\boldsymbol{\mu}_j,\boldsymbol{\Sigma}_j)}.
\label{eq:gmm_responsibility}
\end{equation}
Thus, $\gamma_{ik}\in[0,1]$ is the posterior responsibility of component
$k$ for transformer $i$, and $\sum_{k=1}^{K}\gamma_{ik}=1$.
The hard cluster label used by the forecasting stage is then
\begin{equation}
c_i=\arg\max_{k\in\{1,\ldots,K\}}\gamma_{ik}.
\label{eq:gmm_assignment}
\end{equation}
This yields the disjoint device groups
\begin{equation}
\mathcal{C}_k=\{i\mid c_i=k\},
\qquad k=1,\ldots,K.
\end{equation}
Because $c_i$ is determined in the learned representation space, devices from different administrative regions may be assigned to the same cluster when their temporal load behavior is similar, while devices in the same region may be separated when their patterns differ.

\subsubsection{Cluster Prototype Construction}
The mean of each fitted GMM component is retained conceptually as the prototype of the corresponding load-pattern group,
\begin{equation}
\mathbf{p}_k=\boldsymbol{\mu}_k.
\label{eq:cluster_prototype}
\end{equation}
These prototype vectors provide a compact geometric summary of the
fitted components. The offline labels of the original transformer population
are determined by the GMM posterior responsibilities in
Eq.~\eqref{eq:gmm_responsibility}, not by Euclidean distance to the means.
For the optional lightweight deployment rule described in
Section~\ref{subsec:new_device_assignment}, the retained means are used as
reference vectors for approximate nearest-prototype assignment.

\subsection{Representative Transformer Sampling}
\label{subsec:representative_sampling}

\subsubsection{Cluster-Wise Representative Selection}
Training every transformer independently would remove the principal scalability advantage of clustering. Instead, after the population has been partitioned according to temporal behavior, only a bounded number of devices are used to train each cluster model. For cluster $k$, let $m$ denote the maximum number of training transformers and define
\begin{equation}
m_k=\min(m,|\mathcal{C}_k|).
\end{equation}
A subset
\begin{equation}
\mathcal{S}_k\subseteq\mathcal{C}_k,
\qquad |\mathcal{S}_k|=m_k,
\label{eq:sample_set}
\end{equation}
is sampled uniformly without replacement using a fixed random seed. The sampling is therefore cluster constrained: randomness is used only after the devices have been organized into behaviorally coherent groups. The term ``representative'' in this study refers to this cluster-level coverage rather than to selecting only the devices nearest to a cluster centroid.

\subsubsection{Construction of Cluster Training Sets}
Each sampled transformer contributes multiple supervised
input--target windows. Let $\mathcal{I}^{\mathrm{tr}}_i$ denote the set of
valid training-period forecast origins for transformer $i$, and define
\begin{equation}
\mathcal{D}_i
=
\left\{
\left(\mathbf{X}_{i,t},\mathbf{y}_{i,t}\right)
\,\middle|\,
t\in\mathcal{I}^{\mathrm{tr}}_i
\right\}.
\label{eq:device_dataset}
\end{equation}
The cluster-level supervised training set is then
\begin{equation}
\mathcal{D}_k
=\bigcup_{i\in\mathcal{S}_k}\mathcal{D}_i.
\label{eq:cluster_dataset}
\end{equation}
One PatchTST model $f_k$ is optimized using $\mathcal{D}_k$. The forecasting-side data scaler for cluster $k$ is fitted only on the selected training transformers $\mathcal{S}_k$ and is then kept fixed when the model is applied to the remaining devices in $\mathcal{C}_k$. This prevents information from unsampled test devices from entering the supervised training transformation and ensures that both sampled and unsampled transformers are evaluated under the same cluster-specific model and scaling rule.

\subsection{Cluster-Level Load Forecasting with PatchTST}
\label{subsec:cluster_patchtst}

\begin{figure}[htbp]
    \centering
    \includegraphics[
        width=\textwidth
    ]{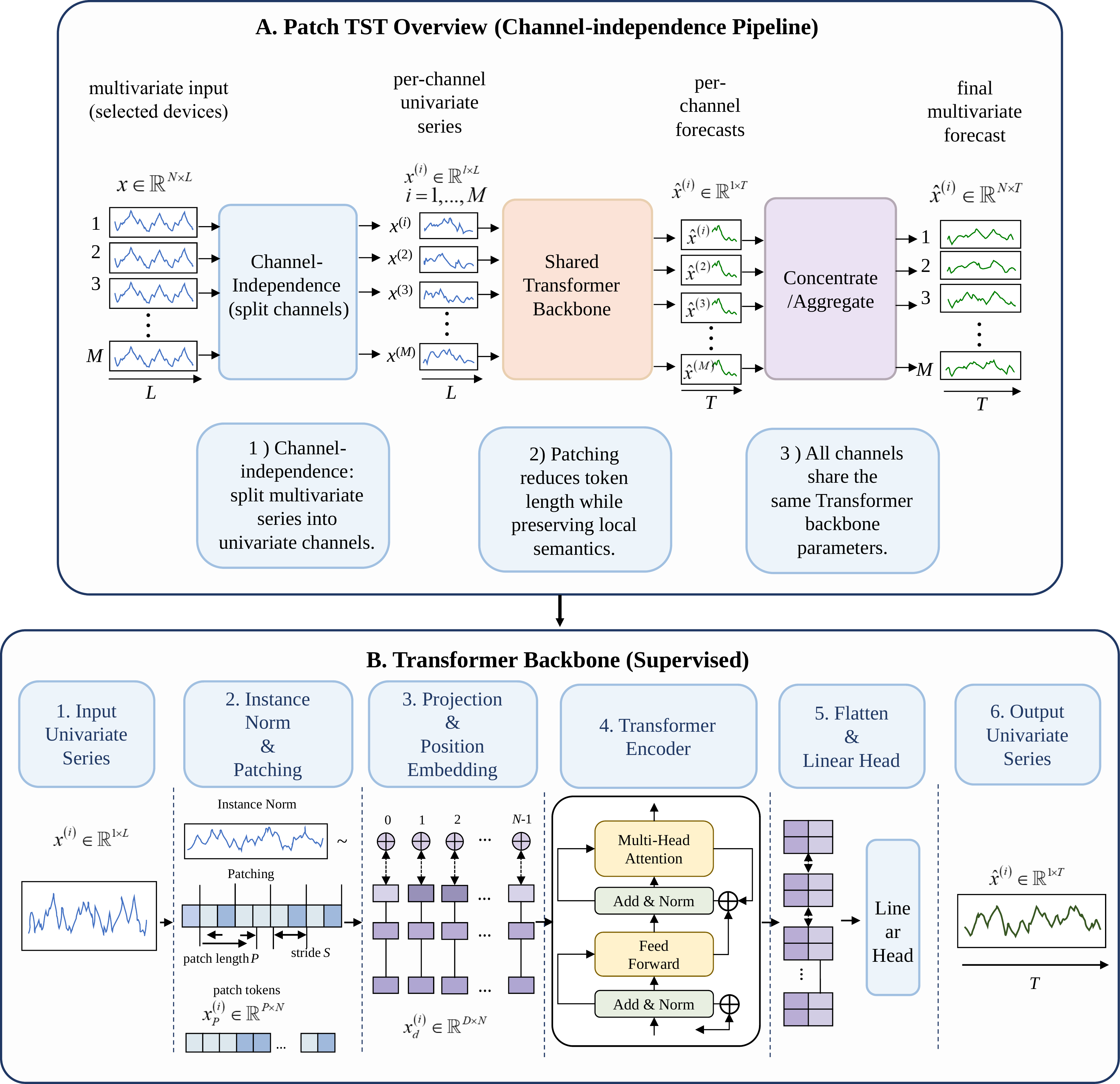}
    \caption{Detailed architecture of the PatchTST forecasting model.
    The figure illustrates the channel-independent forecasting pipeline and
    the internal Transformer-based forecasting backbone.}
    \label{fig:patchtst_architecture}
\end{figure}

The internal forecasting architecture is illustrated in
Fig.~\ref{fig:patchtst_architecture}. The upper-level pipeline shows how the
input variables are processed through the channel-independent mechanism,
whereas the detailed backbone illustrates the patching, embedding,
Transformer encoding, and forecasting stages used by PatchTST.

\FloatBarrier

\subsubsection{Cluster-Level Forecasting Input}
The supervised forecasting stage uses PatchTST as the forecasting backbone \cite{nie2023patchtst}. A critical distinction is that the transformer identifier is \emph{not} treated as a PatchTST channel. Instead, each transformer independently contributes multivariate forecasting samples of the form
\begin{equation}
\mathbf{X}_{i,t}\in\mathbb{R}^{L\times F},
\qquad i\in\mathcal{S}_k,\quad t\in\mathcal{I}^{\mathrm{tr}}_i,
\end{equation}
and all samples generated by the selected transformers of cluster $k$ are pooled into $\mathcal{D}_k$. PatchTST channel independence is applied to the feature channels within each transformer's input sequence. Consequently, the number of transformers in a cluster does not determine the dimensionality of the model input, which allows the same trained cluster model to be evaluated on an unsampled transformer individually.

The implementation follows a multivariate-input, univariate-output setting: multiple numeric input variables may be supplied to the model, while the transformer load is treated as the forecasting target. This organization allows auxiliary variables to be used when available without coupling the architecture to a fixed number of devices.

\subsubsection{Channel-Independent Patching}
For a single feature channel, an input sequence of length $L$ is divided into overlapping patches of length $P$ and stride $S$. Let
$\delta_{\mathrm{pad}}\in\{0,1\}$ indicate whether one additional end patch is
introduced by end padding. The resulting number of patch tokens is
\begin{equation}
N_p=
\left\lfloor\frac{L-P}{S}\right\rfloor+1+\delta_{\mathrm{pad}}.
\label{eq:patch_number}
\end{equation}
In the implemented PatchTST configuration, end padding is enabled, so
$\delta_{\mathrm{pad}}=1$. Each patch is linearly projected into a
$d_m$-dimensional token, where $d_m$ is the PatchTST model dimension, and
positional information is added before Transformer encoding. Reversible
instance normalization (RevIN) is also enabled in the forecasting backbone;
its numerical configuration is reported in Section~4. Compared with processing each individual timestamp as a token, patching shortens the effective token sequence while preserving local temporal context.

PatchTST processes the feature channels independently through the temporal encoder while sharing the same encoder parameters. This design avoids explicitly modeling every cross-channel interaction inside self-attention and provides an efficient temporal backbone for the cluster-specific forecasting task.

\subsubsection{Transformer Encoder and Forecasting Head}
For the $q$th feature channel
$q\in\{1,\ldots,F\}$ of forecasting sample $\mathbf{X}_{i,t}$, let
$\mathbf{Z}^{(q)}_{i,t}\in\mathbb{R}^{N_p\times d_m}$ denote the resulting
patch-token sequence. Using $q$ rather than $f$ avoids confusion with the
forecasting function $f_k$. It is processed through stacked Transformer encoder blocks containing multi-head self-attention, feed-forward layers, residual connections, and normalization. Abstractly,
\begin{equation}
\mathbf{H}^{(q)}_{i,t}
=\mathrm{Encoder}_{\boldsymbol{\theta}_k}
\!\left(\mathbf{Z}^{(q)}_{i,t}\right)
\in\mathbb{R}^{N_p\times d_m}.
\label{eq:patchtst_encoder}
\end{equation}
where $\boldsymbol{\theta}_k$ denotes the parameters of the PatchTST model associated with cluster $k$. The encoded patch features are flattened and mapped through the forecasting head to obtain the future target sequence. The model is trained using the MSE objective in Eq.~\eqref{eq:mse_loss}.

Because the same $\boldsymbol{\theta}_k$ is learned from windows originating from several transformers in $\mathcal{S}_k$, the model captures temporal regularities shared within the cluster rather than memorizing a single device trajectory. At the same time, separate parameter sets $\{\boldsymbol{\theta}_1,\ldots,\boldsymbol{\theta}_K\}$ preserve heterogeneity between clusters.

\subsubsection{Reconstruction of Device-Level Forecasts}
After cluster-level model training, inference is performed transformer by transformer. For transformer $i\in\mathcal{C}_k$, its own input window is transformed using the scaler fitted from $\mathcal{S}_k$ and passed to the saved checkpoint of $f_k$. Predictions are then mapped back to the original load scale. At a common forecasting origin $t$, stacking the $T$-step
device-level forecasts for all $N$ transformers gives
\begin{equation}
\hat{\mathbf{Y}}_t
=
[\hat{\mathbf{y}}_{1,t}^{\mathsf{T}},\ldots,
\hat{\mathbf{y}}_{N,t}^{\mathsf{T}}]^{\mathsf{T}}
\in\mathbb{R}^{N\times T}.
\label{eq:population_forecast}
\end{equation}
Thus, clustering changes the organization and quantity of forecasting models but does not reduce the spatial resolution of the final prediction: forecasts remain available for every individual transformer.

\subsection{Model Assignment for New or Reconfigured Transformers}
\label{subsec:new_device_assignment}

\subsubsection{Representation of an Incoming Transformer}
The learned device organization can also be reused when the forecasting population changes. Let $\mathbf{x}^{(r)}_{*}$ denote the recent historical load sequence of a newly introduced transformer, or of an existing transformer whose operating pattern has changed after network reconfiguration or load transfer. Its mean
$\mu_*^{(r)}$ and standard deviation $\sigma_*^{(r)}$ are computed from the
available representation history using the same within-device rule as
Eq.~\eqref{eq:device_standardization}, producing
$\tilde{\mathbf{x}}_*^{(r)}$. The standardized sequence is then encoded using
the same trained full-series T-Rep operator and $\ell_2$ normalization:
\begin{equation}
\mathbf{e}_{*}
=g_{\phi}^{\mathrm{FS}}(\tilde{\mathbf{x}}^{(r)}_{*}),
\qquad
\mathbf{z}_{*}=\frac{\mathbf{e}_{*}}{\|\mathbf{e}_{*}\|_2}.
\label{eq:new_device_representation}
\end{equation}
No retraining of the representation model is required for this encoding operation.

\subsubsection{Prototype-Based Cluster Assignment}
For the lightweight deployment extension considered here, only the
GMM component means are retained as prototype vectors
$\{\mathbf{p}_k\}_{k=1}^{K}$, with
$\mathbf{p}_k=\boldsymbol{\mu}_k$. An incoming transformer is assigned to the
nearest prototype in the normalized representation space:
\begin{equation}
c_{*}
=\arg\min_{k\in\{1,\ldots,K\}}
\|\mathbf{z}_{*}-\mathbf{p}_k\|_2.
\label{eq:new_device_assignment}
\end{equation}
This nearest-prototype rule should be distinguished from the posterior
GMM assignment in Eq.~\eqref{eq:gmm_assignment}, which is used for the
original offline clustering. The prototype rule is a lightweight deployment
approximation when only component means are retained; it is not used to
produce the reported offline clustering results.

\subsubsection{Forecasting without Model Retraining}
Once $c_{*}$ is obtained, the incoming device reuses the forecasting model associated with that cluster:
\begin{equation}
\hat{\mathbf{y}}_{*,t}
=f_{c_{*}}(\mathbf{X}_{*,t};\boldsymbol{\theta}_{c_{*}}).
\end{equation}
Accordingly, population changes do not necessarily require an additional device-specific PatchTST. The assignment can also be recomputed periodically when sufficient recent history becomes available, allowing the device-to-model mapping to adapt to persistent pattern changes. This mechanism is a deployment capability of the proposed framework; its role is to extend the cluster-level organization to devices outside the original supervised training subset.

\subsection{Rated-Capacity Warning Formulation}
\label{subsec:warning_formulation}

\subsubsection{Continuous Forecast to Warning State}
In addition to continuous load prediction, the framework converts forecasting trajectories into an engineering-oriented high-load warning state. Let $C_i$ denote the rated capacity of transformer $i$. For a rated-capacity ratio
$\alpha\in(0,1)$, define the device-specific threshold as
\begin{equation}
\tau_i(\alpha)=\alpha C_i.
\label{eq:warning_threshold}
\end{equation}
The operational pre-warning rule considered in the application uses
$\alpha_{\mathrm{op}}=0.8$, whereas the statistical warning evaluation reported
in the present experiments uses $\alpha_{\mathrm{eval}}=0.5$, as specified in
Section~4. Keeping $\alpha$ explicit in the formulation distinguishes the
operational reference threshold from the threshold used to compute the
reported classification metric. For any chosen $\alpha$, the observed and
predicted pointwise warning states are
\begin{equation}
s_{i,t}^{(\alpha)}
=\mathbb{I}\!\left(y_{i,t}\ge\tau_i(\alpha)\right),
\qquad
\hat{s}_{i,t}^{(\alpha)}
=\mathbb{I}\!\left(\hat{y}_{i,t}\ge\tau_i(\alpha)\right),
\label{eq:pointwise_warning}
\end{equation}
where $\mathbb{I}(\cdot)$ is the indicator function. This transformation allows a forecasting model to be evaluated not only by its average numerical error but also by whether it recognizes periods approaching an operationally critical loading region.

\subsubsection{Pointwise Warning Evaluation Unit}
Consistent with the experimental evaluation in Section~4, each 15-min
timestamp is treated as one warning-evaluation instance. Using
$\alpha_{\mathrm{eval}}=0.5$, the pair
$\bigl(s_{i,t}^{(\alpha_{\mathrm{eval}})},
\hat{s}_{i,t}^{(\alpha_{\mathrm{eval}})}\bigr)$ determines whether that
timestamp is a true positive, true negative, false positive, or false
negative. Warning accuracy and the corresponding confusion-matrix notation
are formally defined in Section~4. No device-day aggregation is applied to the
reported warning accuracy.

\subsection{Scalability Analysis}
\label{subsec:scalability_analysis}

\subsubsection{Model-Count Reduction}

The scalability benefit of the proposed framework is first reflected in
the number of forecasting models that need to be trained, stored, and
maintained. A fully local strategy requires
\begin{equation}
N_{\mathrm{model}}^{\mathrm{local}}=N,
\end{equation}
whereas a single global model requires
\begin{equation}
N_{\mathrm{model}}^{\mathrm{global}}=1.
\end{equation}
The proposed framework uses
\begin{equation}
N_{\mathrm{model}}^{\mathrm{proposed}}=K,
\qquad 1<K<N,
\end{equation}
where $K$ is the number of behavior-based clusters and is typically much
smaller than the number of transformers $N$ in a large-population setting.

Relative to the one-model-per-transformer strategy, the reduction in
forecasting-model quantity is
\begin{equation}
R_{\mathrm{model}}
=1-\frac{K}{N}.
\label{eq:model_reduction}
\end{equation}
This ratio quantifies the reduction in independently stored and maintained
forecasting parameter sets. It does not imply that the proposed framework has
the same computational characteristics as a single global model, because each
of the $K$ cluster-specific models is trained separately. The corresponding
model-count reduction for the experimental transformer population is reported
in Section~5.

\subsubsection{Training-Device and Deployment Scale}

Representative sampling reduces the number of transformers contributing
to the supervised training of each cluster-specific PatchTST model. Let
$m_k=|\mathcal{S}_k|$ denote the number of sampled transformers in cluster
$k$, and let $m$ denote the maximum number of transformers sampled from one
cluster. Since $m_k\leq m$, the total number of transformers contributing to
supervised forecasting-model training satisfies
\begin{equation}
N_{\mathrm{train\mbox{-}dev}}
=\sum_{k=1}^{K}m_k
\leq Km.
\label{eq:training_device_bound}
\end{equation}

Importantly, this bound applies only to the supervised PatchTST training
stage. The preceding T-Rep representation learning and GMM clustering stages
still use the transformer population required for model construction.
Therefore, Eq.~\eqref{eq:training_device_bound} should be interpreted as a
measure of supervised training scale rather than as a direct estimate of the
end-to-end computational complexity of the complete framework.

When $K\ll N$ and the per-cluster sampling cap $m$ is small relative to
the full transformer population, only a limited subset of devices contributes
forecasting windows to PatchTST training. The resulting runtime saving,
however, also depends on factors such as the number of sliding windows,
cluster sizes, early-stopping behavior, batch processing, and hardware
utilization. For this reason, the present study does not infer wall-clock
training time, inference latency, memory consumption, or computational energy
use directly from Eq.~\eqref{eq:training_device_bound}.

At deployment, each transformer invokes only one of the $K$ stored PatchTST
models.
For transformer $i$, the cluster label $c_i$ determines the forecasting
model $f_{c_i}$ to be used, while the prediction itself remains based on that
transformer's own input window. Consequently, the number of device-level
forecasts still scales with $N$, as required for fine-grained operation,
whereas forecasting-model storage, version management, updating, and
retraining are organized at the cluster level rather than at the
individual-device level. Section~5 reports the resulting model count and
supervised training-device scale for the experimental setting.


\section{Experimental setup}
\subsection{Dataset Description}
\subsubsection{Data Sources and Study Population}
This study uses measured operational data from an original population of
1,065 distribution transformers collected over a four-year period in an
undisclosed region of China. After data cleaning and quality screening,
998 transformers with complete and usable records from May 1, 2024, to
August 31, 2025, were retained for the experiments in this study.
The location of the study area cannot be disclosed under the applicable data
confidentiality agreement. Each transformer is identified by a unique device
identifier. The forecasting target is transformer load, measured in kVA.
The dataset also records each transformer's rated capacity, likewise measured
in kVA, which is used to define the high-load warning threshold.
Missing load values and anomalous spikes are handled according to Subsection 4.2.

\subsubsection{Exploratory Analysis of Transformer Load Characteristics}

To further investigate load heterogeneity across distribution transformers, an exploratory data analysis was conducted to examine the statistical characteristics of transformer load distributions. Table 3 summarizes the main characteristics of the dataset, including the number of transformers, observation period, sampling resolution, forecasting variables, and data partitioning.

\begin{table}[pos=htbp]
\centering
\caption{Summary of the distribution transformer dataset and experimental configuration.}
\label{tab:dataset_overview}
\small
\renewcommand{\arraystretch}{1.15}
\begin{tabular}{@{}p{0.34\linewidth}p{0.61\linewidth}@{}}
\toprule
\textbf{Characteristic} & \textbf{Description} \\
\midrule
Study area & Undisclosed region in China \\

Number of transformers & 998 \\

Observation period & May 1, 2024 -- August 31, 2025 \\

Load measurement & Transformer apparent power (kVA) \\

Rated capacity & Device-specific rated capacity (kVA) \\

Load sampling interval & 15 minutes \\

Observations per day & 96 per transformer \\

Meteorological data source & China Meteorological Data Service Centre \\



Historical input window & 96 time steps (24 hours) \\

Forecasting horizon & 96 time steps (24 hours) \\

Data partition & Training 80\%, validation 10\%, test 10\% \\

Data splitting strategy & Chronological \\
\bottomrule
\end{tabular}
\end{table}

\begin{figure}
    \centering
    \includegraphics[width=0.9\linewidth]{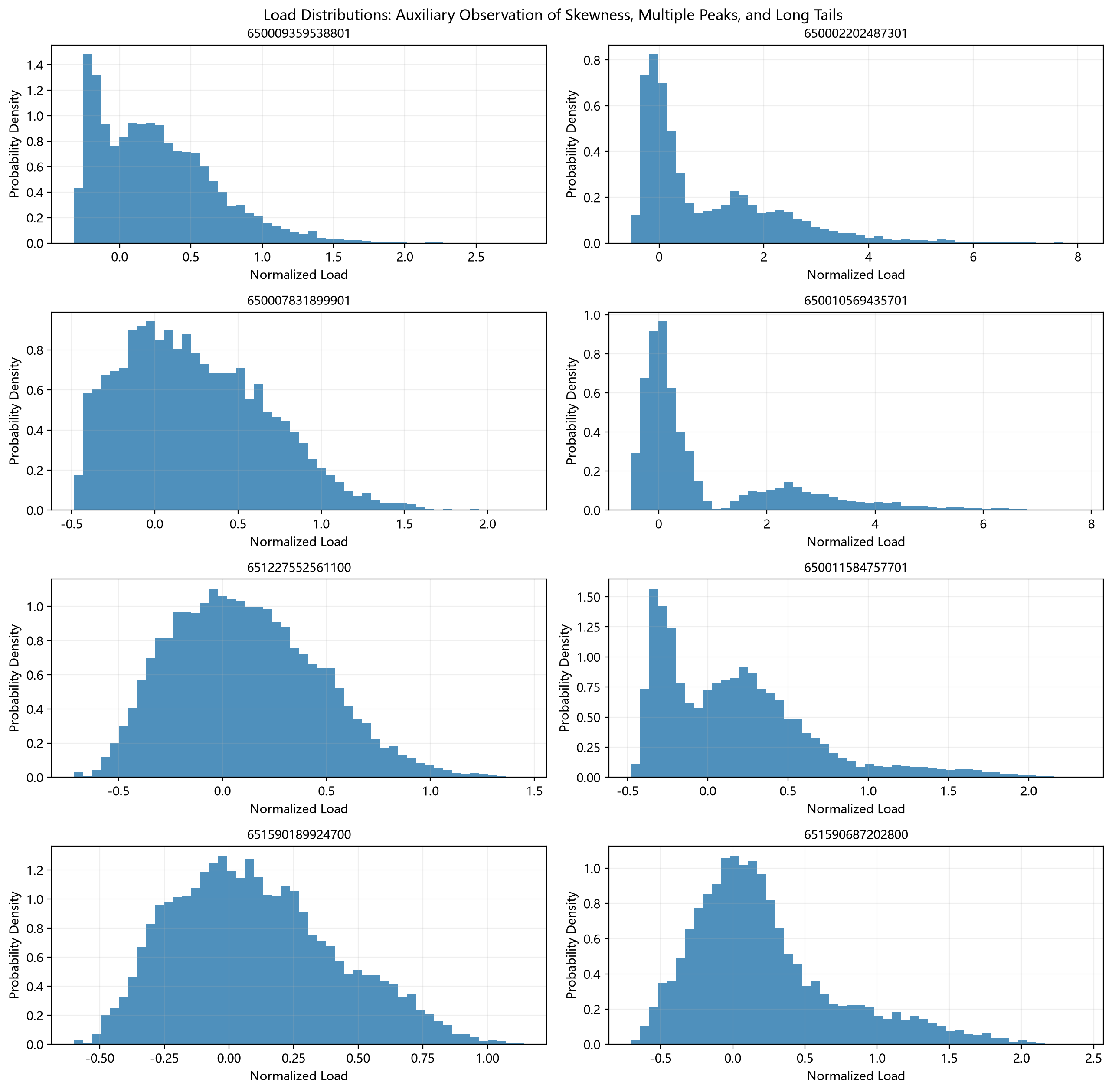}
    \caption{Normalized load distributions of eight selected distribution transformers, illustrating variations in distribution shape, skewness, multimodality, and tail behavior.}
    \label{fig:load_distributions}
\end{figure}

\begin{figure}[p]
    \centering

    \begin{subfigure}{\linewidth}
        \centering
        \includegraphics[
            width=\linewidth,
            height=0.80\textheight,
            keepaspectratio
        ]{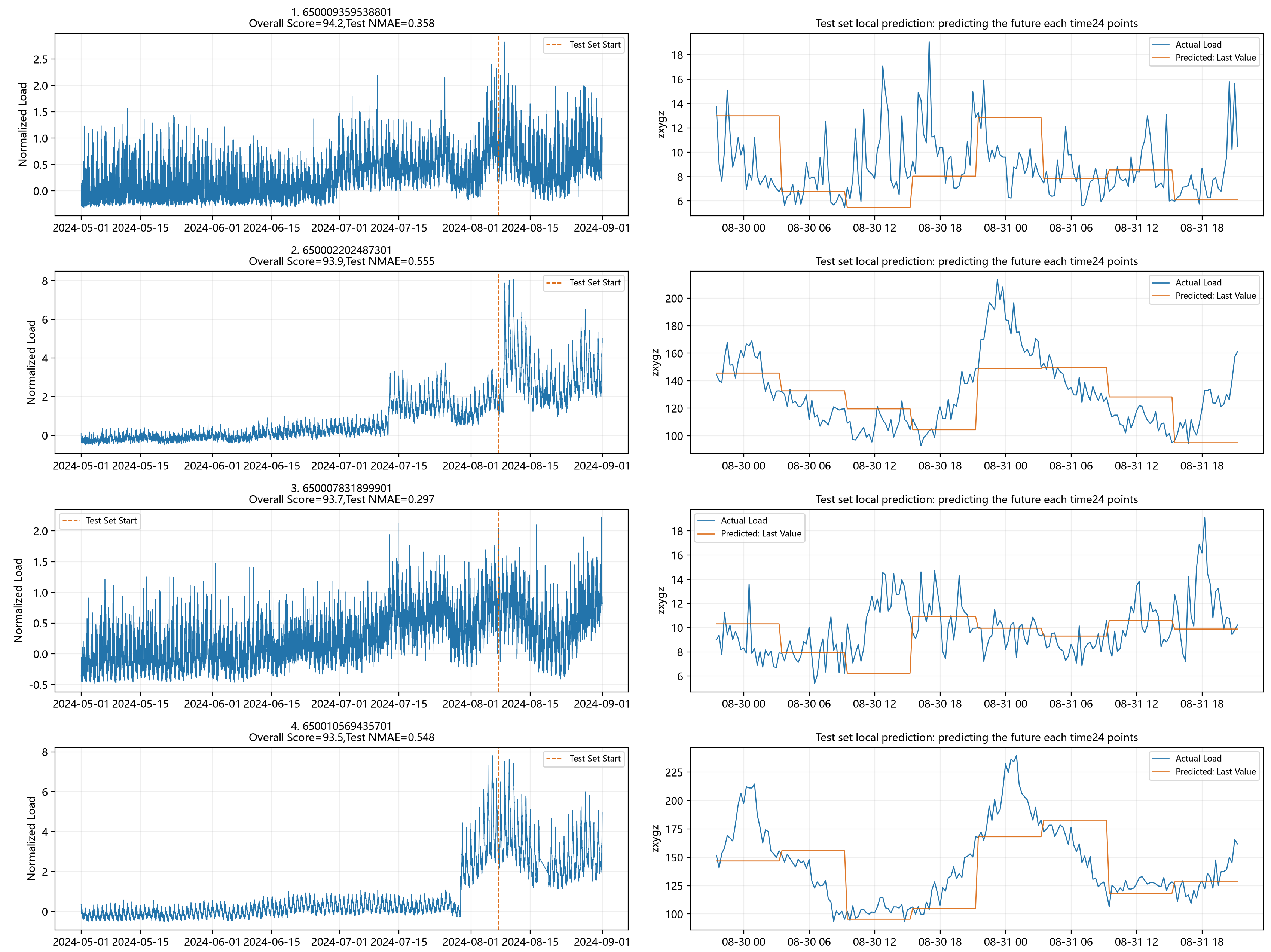}
        \caption{Load profiles of Transformers 1--4.}
        \label{fig:selected_devices_1}
    \end{subfigure}

    \caption{Temporal variations in normalized electricity
    consumption across eight representative load profiles
    (Part 1 of 2).}
    \label{fig:selected_devices}
\end{figure}

\begin{figure}[p]
    \ContinuedFloat
    \centering

    \setcounter{subfigure}{1}

    \begin{subfigure}{\linewidth}
        \centering
        \includegraphics[
            width=\linewidth,
            height=0.80\textheight,
            keepaspectratio
        ]{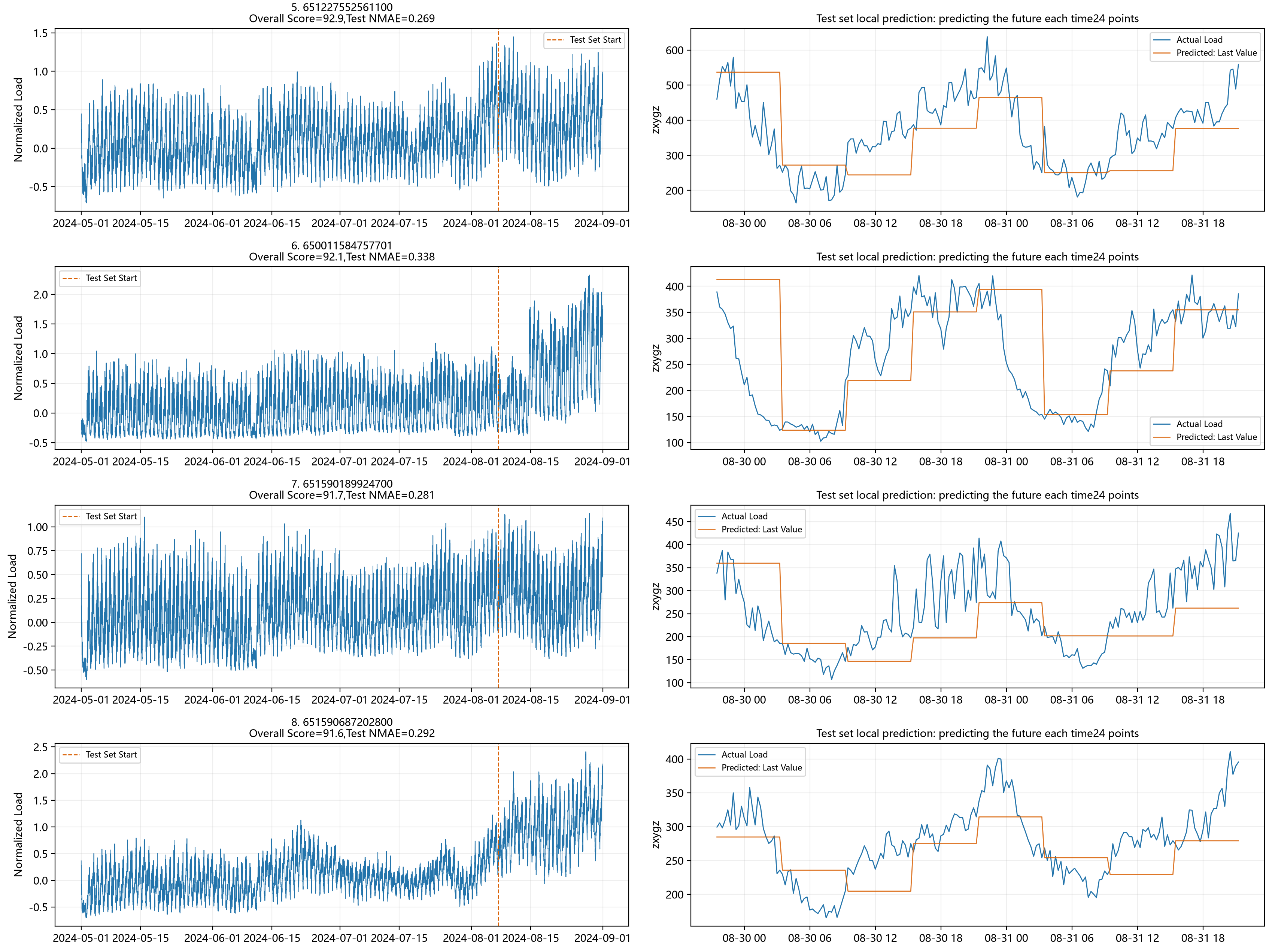}
        \caption{Load profiles of Transformers 5--8.}
        \label{fig:selected_devices_2}
    \end{subfigure}

    \caption[]{Temporal variations in normalized electricity
    consumption across eight representative load profiles
    (continued).}
\end{figure}

\clearpage

Furthermore, Figure 3 presents the normalized load distributions of eight transformers randomly selected following an auxiliary load-complexity analysis. Specifically, historical load observations from each transformer were normalized and visualized using probability density histograms, where the horizontal axis represents normalized load and the vertical axis indicates probability density. Considerable differences can be observed across transformers. Some devices exhibit relatively concentrated unimodal distributions, whereas others show pronounced skewness, multimodal patterns, or long-tailed distributions. Moreover, Figure 4 illustrates the normalized power consumption variations of eight representative load profiles over a one-year observation period. By converting different load sequences into per-unit values, their temporal variation characteristics can be compared on a similar numerical scale. It can be observed that the different load profiles exhibit significant differences in long-term trends, periodicity, and fluctuation amplitude. Some loads display relatively stable periodic fluctuations, while others show pronounced seasonal variations or shifts in load levels. These observations indicate that transformers differ not only in their load magnitudes but also in the statistical characteristics of their load variations.

\subsection{Data Preprocessing and Dataset Construction}
\subsubsection{Data Cleaning and Missing-Value Treatment}
Load records are first sorted by device identifier and timestamp and then examined for missing observations and anomalous spikes. An isolated missing value is imputed only when valid observations are available at both immediately adjacent timestamps. The imputed value is the arithmetic mean of the two adjacent load values. If $L_{i,t}$ denotes the load of transformer $i$ at time $t$, the imputation is
\begin{equation}
    \widehat{L}_{i,t}=\frac{L_{i,t-1}+L_{i,t+1}}{2}.
    \label{eq:single-point-imputation}
\end{equation}
Runs of two or more consecutive missing observations are not interpolated. These runs, together with forecasting windows that cross them, are excluded from subsequent sample construction.

To identify isolated spikes that may arise from measurement errors, a local window comprising the four neighboring observations on either side of the candidate point, excluding the candidate itself, is proposed. Let $m_{i,t}$ and $\operatorname{MAD}_{i,t}$ denote the median and median absolute deviation within this window, and define $s_{i,t}=1.4826\operatorname{MAD}_{i,t}$. A point is classified as an anomalous spike only if it exceeds both the local median and its immediately adjacent observations, with both excesses greater than $3s_{i,t}$. This criterion targets isolated upward spikes; sustained high-load periods are not removed solely by this rule. For an isolated anomalous point, the original load value is removed and replaced using Eq.~\eqref{eq:single-point-imputation}, provided that both adjacent observations are valid. Consecutive anomalous points are treated as a missing run.

\subsubsection{Normalization and Forecasting Sample Construction}
T-Rep and PatchTST use normalization procedures suited to their respective training tasks. For T-Rep, the mean and standard deviation of each transformer's load series are estimated from its training period and used for within-device normalization. This emphasizes the temporal shape of the load profile. For cluster-level PatchTST, the scaler is fitted only on training-period observations from the representative transformers selected for that cluster. The resulting parameters are then applied to the training, validation, and test observations of devices in the same cluster. Neither normalization procedure estimates its parameters from the validation or test period.

Forecasting samples are constructed separately for each transformer. Each sample consists of 96 consecutive historical observations followed by 96 consecutive target observations, with a sliding-window stride of one observation. Windows do not cross device boundaries or the boundaries between the training, validation, and test sets. 

\subsubsection{Training, Validation, and Test Splits}
The data are divided chronologically in an 8:1:1 ratio: the earliest 80\% form the training set, the next 10\% the validation set, and the final 10\% the test set. T-Rep representation learning, Gaussian mixture model (GMM) clustering, and representative-transformer selection use only training-period data. PatchTST parameters are updated on the training set, while the validation set is used for early stopping, checkpoint selection, and selection of the GMM cluster number according to mean validation MAE across three experimental runs. The test set is reserved for final performance evaluation; results for the remaining candidate cluster numbers are reported only as a sensitivity analysis and do not influence the selected configuration. At each stage, normalization parameters are estimated only from the training data available to that stage.

\subsection{Experimental Configuration}
\subsubsection{Forecast Inputs and Prediction Horizon}
The models use 96 consecutive 15-minute observations as historical input and directly forecast the following 96 observations. Thus, each task uses the preceding 24 hours to predict transformer load over the next 24 hours. The input is multivariate, and device load is the forecasting target. The data loader constructs sliding windows within each transformer's time series; windows do not cross device or dataset-split boundaries. In the training code, numeric columns other than the timestamp, device identifier, and split label are used as inputs, with the load target placed in the final channel.

\subsubsection{T-Rep and GMM Configuration}
Load series are normalized within each device using their training-period means and standard deviations before being passed to T-Rep. T-Rep uses a learnable sinusoidal time embedding and produces 128-dimensional representations. It is trained for 40 epochs with a batch size of 8, a learning rate of $10^{-3}$, a maximum training-sequence length of 3,000 time points, and a random seed of 0. The optimizer is AdamW. Four self-supervised losses---instance contrast, temporal contrast, time-embedding divergence prediction, and time-embedding-conditioned forecasting---are each assigned a weight of 0.25. Both T-Rep training and device-representation extraction use training-period load series only, preventing validation- or test-period data from entering the clustering stage. After training, one device-level representation vector is extracted for each transformer's training-period series.

The T-Rep representations are clustered using a GMM with
four candidate clustering granularities, $K=3$, $5$, $7$, and $10$.
The implementation uses diagonal covariance matrices, covariance
regularization of $10^{-6}$, a maximum of 300 expectation--maximization
iterations, 10 initializations, and a GMM random state of 42.

The cluster number $K$ is selected using only the validation set. For each
candidate value, representative-device sampling and forecasting-model training
are repeated with three random seeds, and the mean validation MAE is used as
the predefined selection criterion. The selected value is fixed before final
test-set evaluation, while the remaining candidate values are retained only
for sensitivity analysis.

\subsubsection{PatchTST and Training Configuration}
Up to 20 representative transformers are randomly selected from each cluster to train one forecasting model, which is then shared by all devices in that cluster at inference time. The same per-cluster sampling limit is used for the experiments with 3, 5, 7, and 10 clusters. For each cluster-count setting, the representative transformers are randomly selected on three separate occasions, and a forecasting model is trained for each selection. The reported result is the mean of the three experimental runs.

PatchTST uses a patch length of 16 and a stride of 8. Its encoder has two layers and eight attention heads, with a model dimension of 512 and a feed-forward dimension of 2,048. Reversible instance normalization (RevIN) is enabled. Dropout in the model body and fully connected layer is 0.05. Training uses Adam and mean squared error (MSE) loss, with a batch size of 128, an initial learning rate of $10^{-4}$, and a maximum of 50 epochs. If validation loss fails to improve for 10 consecutive epochs, training stops and the checkpoint with the lowest validation loss is restored. The learning rate remains at its initial value for the first three epochs and is subsequently multiplied by 0.9 after each epoch.

For each cluster, normalization parameters are fitted only to training-period observations from its selected representative transformers and are then used to transform that cluster's training and inference data. The experiments were conducted using an NVIDIA GeForce RTX 3070 GPU and Python 3.12.

\subsection{Comparison Methods}
\subsubsection{Forecasting-Model Comparisons}
The forecasting comparisons include seven models: iTransformer, FiLM, TimesNet, Pyraformer, FEDformer, DLinear, and Informer. A PatchTST model trained directly without T-Rep representation learning or GMM clustering serves as an additional baseline, denoted \texttt{baseline} in the results table. All models use the same 998-device dataset, the chronological 8:1:1 split, and the 96-input/96-output forecasting task.

Because training a single forecasting model on every device's data is impractical under the intended deployment conditions, each of the seven comparison models and the direct PatchTST baseline is trained on a separate random sample of 300 devices, using only their training-period observations. The sampled groups of 300 devices are not identical across models. For the proposed method, T-Rep representations and GMM clusters are first obtained from the training-period series of all 998 devices; a cluster-level PatchTST model is then trained on up to 20 randomly selected devices per cluster. Every model is evaluated on the test-period data of all 998 devices. Consequently, performance differences should be interpreted in light of both the number and composition of devices used to train each forecasting model.

\subsubsection{Selection and Sensitivity Analysis of the GMM Cluster Number}
Four candidate GMM cluster numbers, $K=3$, $5$, $7$, and $10$, are
considered while all remaining experimental settings are kept unchanged. For
each candidate value, representative-device sampling and forecasting-model
training are conducted with three random seeds, and the resulting validation
MAE values are averaged across the three runs. The cluster number is selected
before any test-set comparison using the predefined criterion of minimum mean
validation MAE.

\begin{table}[pos=htbp]
\centering
\caption{Validation-set MAE for candidate GMM cluster numbers used in cluster-number selection.}
\label{tab:gmm_validation_selection}
\begin{tabular}{cr}
\toprule
$K$ & Validation MAE \\
\midrule
3  & 22.3216 \\
5  & 20.7749 \\
\textbf{7}  & \textbf{18.2854} \\
10 & 20.3691 \\
\bottomrule
\end{tabular}
\end{table}
\FloatBarrier

As shown in Table~\ref{tab:gmm_validation_selection}, $K=7$ yields the
lowest mean validation MAE of 18.2854 and is therefore selected as the ClusLoad
configuration for the primary comparison with other forecasting methods. This
value is fixed before final test-set evaluation. After model selection, all four
candidate values are additionally evaluated on the test set to examine the
sensitivity of ClusLoad to clustering granularity. These test-set results are
reported in Section~\ref{sec:cluster_sensitivity} and do not influence the
selection of $K$.

\subsubsection{Random-Grouping Ablation}

To isolate the contribution of behavior-based transformer organization, a
random-grouping ablation is constructed by replacing the T-Rep--GMM
grouping stage with random device assignment. For each evaluated value of
$K$, the number of transformers assigned to each random group is kept
identical to the corresponding cluster size produced by T-Rep and GMM.
The subsequent representative-device sampling, PatchTST architecture,
training procedure, and evaluation protocol remain unchanged.

This design controls the number and sizes of forecasting groups and therefore
allows the effect of behavior-based clustering to be evaluated independently
of changes in model quantity or supervised training scale.

\subsection{Evaluation Metrics}
\subsubsection{Continuous Load-Forecasting Metrics}
Continuous load forecasts are evaluated using mean absolute error (MAE), mean squared error (MSE), root mean squared error (RMSE), mean absolute percentage error (MAPE), and mean squared percentage error (MSPE). Before computing these metrics, predictions and observations are transformed back to the original load scale in kVA. All 96 predicted points in each test window are included. MAE and RMSE are measured in kVA, MSE in kVA$^2$, and MAPE and MSPE are dimensionless.

Each metric is computed separately for each transformer and then averaged arithmetically over the 998 transformers, giving every device equal weight in the overall result. If $M_i$ is a metric for device $i$, the reported device-level mean is
\begin{equation}
    \overline{M}=\frac{1}{998}\sum_{i=1}^{998}M_i.
    \label{eq:device-macro-average}
\end{equation}
For MAPE and MSPE, $\varepsilon=10^{-3}$ is added to the denominator to stabilize the calculation when observed load is zero or near zero. For GMM experiments repeated with three random seeds, the three device-averaged values are then averaged arithmetically. Because RMSE is square-rooted at the device level before averaging across devices, the square of the overall RMSE is not generally equal to the overall MSE.

\subsubsection{High-Load Warning Evaluation}
This study addresses high-load warning for distribution transformers. Continuous forecasting errors measure how closely a model estimates the numerical load, whereas warning outcomes determine whether operators can respond in time. Warning accuracy is therefore the principal application-oriented metric in this study.

Under the operational rule, a warning is triggered when the predicted load reaches 80\% of the transformer's rated capacity. This threshold provides time and capacity margin for operators to transfer load or adjust power flows before sustained high loading develops into an overload. Timely warnings can reduce the risk of transformer damage and power interruptions. The study area contains many factories, for which interruptions may disrupt production; identifying high-load periods accurately is therefore particularly important in this setting.

For statistical evaluation, however, only a small fraction of observations in the available test period exceed 80\% of rated capacity, resulting in a highly imbalanced warning dataset and an insufficient number of positive instances for stable comparison across forecasting methods. Therefore, a threshold of 50\% of rated capacity is adopted for the comparative warning evaluation, while the 80\% threshold is retained as the practical operational reference rather than being replaced. Each 15-minute time point is treated as one warning-evaluation instance. Let $C_i$ be the rated capacity of transformer $i$. A predicted load of at least $0.5C_i$ is classified as a predicted warning, and an observed load of at least $0.5C_i$ as a true warning. Warning accuracy is defined as the proportion of correctly predicted warning and non-warning instances among all evaluated observations:
\begin{equation}
    \mathrm{Accuracy}=\frac{TP+TN}{TP+TN+FP+FN},
    \label{eq:warning-accuracy}
\end{equation}
where $TP$, $TN$, $FP$, and $FN$ denote true positives, true negatives, false positives, and false negatives, respectively. Continuous forecasting errors are reported alongside warning accuracy to characterize numerical forecasting performance, while warning accuracy is emphasized when discussing practical utility.


\section{Results and Discussion}
\label{sec:results}

\subsection{Overall Forecasting Performance}

\subsubsection{Comparison with Forecasting Models}

Table~\ref{tab:forecasting_comparison} compares the proposed method with
seven representative time-series forecasting models and the direct PatchTST
baseline. The proposed method achieves the lowest MAE, MSE, RMSE, MAPE,
and MSPE among all compared methods, with values of 18.7997, 746.7468,
24.6080, 0.0950, and 0.0125, respectively. It also achieves the highest
warning accuracy of 95.66\%. These results indicate that organizing
transformers according to their learned temporal load patterns before
forecasting provides a clear advantage over directly applying a forecasting
model to randomly sampled or unstructured device populations.

\begin{table}[pos=htbp]
\centering
\caption{Comparison of forecasting and warning performance among different
models. Lower values are better for MAE, MSE, RMSE, MAPE, and MSPE, whereas
higher values are better for warning accuracy. PatchTST denotes the direct
baseline model.}
\label{tab:forecasting_comparison}

\small
\setlength{\tabcolsep}{4pt}
\renewcommand{\arraystretch}{1.08}

\begin{tabular}{lrrrrrr}
\toprule
Method & MAE & MSE & RMSE & MAPE & MSPE & Accuracy(\%) \\
\midrule
iTransformer
    & 24.4000 & 1227.8754 & 31.8579 & 0.1347 & 0.2732 & 87.21 \\
FiLM
    & 22.7600 & 975.4483 & 29.5460 & 0.1344 & 0.5081 & 88.30 \\
TimesNet
    & 22.5500 & 974.5294 & 29.4536 & 0.1165 & 0.0555 & 87.37 \\
Pyraformer
    & 23.4730 & 1048.8007 & 30.5759 & 0.1202 & 0.0552 & 86.64 \\
FEDformer
    & 22.6800 & 969.9426 & 29.4284 & 0.1250 & 0.2111 & 87.78 \\
DLinear
    & 30.3174 & 1812.8382 & 42.5774 & 0.2337 & 0.1316 & 86.10 \\
Informer
    & 29.9800 & 1476.3796 & 38.4237 & 0.1441 & 0.0346 & 83.95 \\
PatchTST$^{*}$
    & 23.4300 & 1053.8578 & 30.4793 & 0.1965
    & 0.1439 & 88.14 \\
\textbf{ClusLoad}$^{**}$
    & \textbf{18.7997}
    & \textbf{746.7468}
    & \textbf{24.6080}
    & \textbf{0.0950}
    & \textbf{0.0125}
    & \textbf{95.66} \\
\bottomrule
\end{tabular}

\vspace{2pt}
\begin{minipage}{0.98\linewidth}
\footnotesize
$^{*}$ PatchTST is used as the direct baseline for evaluating the proposed
cluster-based model-sharing strategy.\\
$^{**}$ ClusLoad denotes the $K=7$ configuration selected using the
validation set and fixed before final test-set evaluation.
\end{minipage}
\end{table}
\FloatBarrier

Compared with the direct PatchTST baseline, the proposed framework reduces
MAE by approximately 19.76\%, MSE by 29.14\%, RMSE by 19.26\%, MAPE by
51.65\%, and MSPE by 91.31\%. The warning accuracy increases from 88.14\% to
95.66\%, corresponding to an improvement of 7.52 percentage points. Because
PatchTST is retained as the forecasting backbone in both settings, this
comparison indicates that the performance gains are closely associated with
the representation--clustering--sampling strategy introduced before
forecasting rather than with a change in forecasting architecture.

A similar advantage is observed when ClusLoad is compared with the strongest
alternative models. Among these methods, TimesNet achieves the lowest MAE and
MAPE, FEDformer achieves the lowest MSE and RMSE, and Informer achieves the
lowest MSPE. Relative to these strongest competing values, ClusLoad reduces
MAE by approximately 16.63\%, MSE by 23.01\%, RMSE by 16.38\%, MAPE by
18.45\%, and MSPE by 63.87\%. The consistent improvements across different
error criteria suggest that the proposed model-sharing strategy improves
forecasting performance broadly rather than optimizing a single metric. These
results are particularly notable because ClusLoad is designed primarily to
address cross-device heterogeneity and deployment scalability rather than to
introduce a new forecasting backbone.

\subsubsection{Relationship between Forecasting Error and Warning Performance}



The forecasting models exhibit different rankings in terms of continuous prediction errors and warning accuracy, indicating that numerical forecasting performance and high-load state identification are not fully interchangeable. For instance, FiLM achieves the highest warning accuracy among the comparison models at 88.30\%, despite not obtaining the lowest continuous forecasting errors. Similarly, TimesNet performs well in terms of MAE and MAPE but achieves a warning accuracy of only 87.37\%. These differences suggest that lower average prediction errors do not necessarily translate into more accurate identification of high-load conditions.

This discrepancy can be understood from the perspectives of load distribution characteristics and threshold-based classification. First, the exploratory analysis in Section 4.1.3 reveals considerable differences in transformer load distributions, including skewed, multimodal, and long-tailed patterns. Such heterogeneity suggests that the distribution of observations near high-load thresholds may vary across devices. Second, warning outcomes depend not only on the magnitude of forecasting errors but also on the positions of predicted and observed loads relative to the warning threshold. When the actual load approaches the threshold, even a relatively small prediction error may lead to an incorrect warning classification. Conversely, larger prediction errors occurring far from the threshold may not change the warning state. Therefore, continuous forecasting errors and warning accuracy provide complementary perspectives on model performance.

The proposed framework achieves favorable performance in both continuous forecasting metrics and warning accuracy, reaching a warning accuracy of 95.66\% under the experimental evaluation threshold. These results demonstrate that the framework can improve overall forecasting accuracy while providing more reliable identification of high-load states. Nevertheless, these observations do not directly explain the internal decision-making mechanisms of the forecasting models. Further model attribution analysis would be required to investigate the contributions of different input features to the forecasting results.


\subsection{Sensitivity to the Number of GMM Clusters}
\label{sec:cluster_sensitivity}

The number of clusters determines the granularity at which heterogeneous
transformers share forecasting models. A small value of $K$ leads to larger
clusters and stronger model sharing, whereas a large $K$ creates more
specialized groups but reduces the amount of representative training data
available to each forecasting model. After $K=7$ was selected on the
validation set using the criterion reported in Table~\ref{tab:gmm_validation_selection},
all four candidate values, $K=3$, $5$, $7$, and $10$, were additionally
evaluated on the test set while retaining the same representation-learning,
sampling, and forecasting procedures. This comparison is used only as a
sensitivity analysis of clustering granularity.

\begin{table}[pos=htbp]
\centering
\caption{Test-set sensitivity of forecasting performance to the number of GMM clusters.}
\label{tab:gmm_cluster_sensitivity}
\begin{tabular}{crrrrr}
\toprule
$K$ & MAE & MSE & RMSE & MAPE & MSPE \\
\midrule
3
    & 21.1800 & 843.5056 & 27.5411 & 0.1866 & 0.7467 \\
5
    & 20.6024 & 869.8537 & 27.0686 & 0.1716 & 0.6872 \\
\textbf{7}
    & \textbf{18.7997}
    & \textbf{746.7468}
    & \textbf{24.6080}
    & \textbf{0.0950}
    & \textbf{0.0125} \\
10
    & 19.9582 & 762.6748 & 26.1099 & 0.2288 & 4.8760 \\
\bottomrule
\end{tabular}
\end{table}
\FloatBarrier

\begin{figure}
    \centering
    \includegraphics[width=1.0\linewidth]{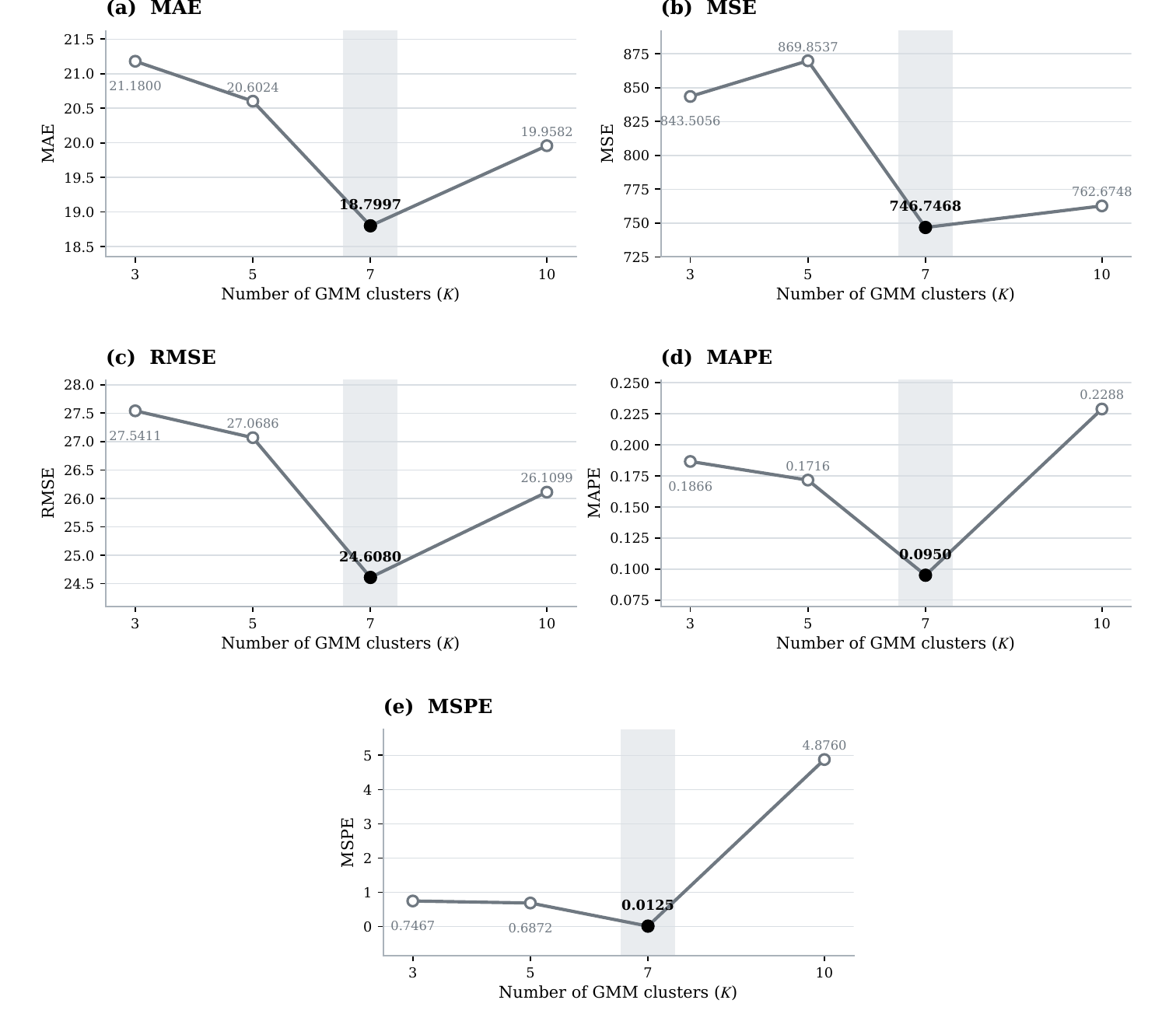}
    \caption{Test-set forecasting performance under different numbers of GMM clusters ($K=3$, $5$, $7$, and $10$), evaluated using MAE, MSE, RMSE, MAPE, and MSPE.}
    \label{fig:clustering}
\end{figure}

\begin{figure}
    \centering
    \includegraphics[width=1.0\linewidth]{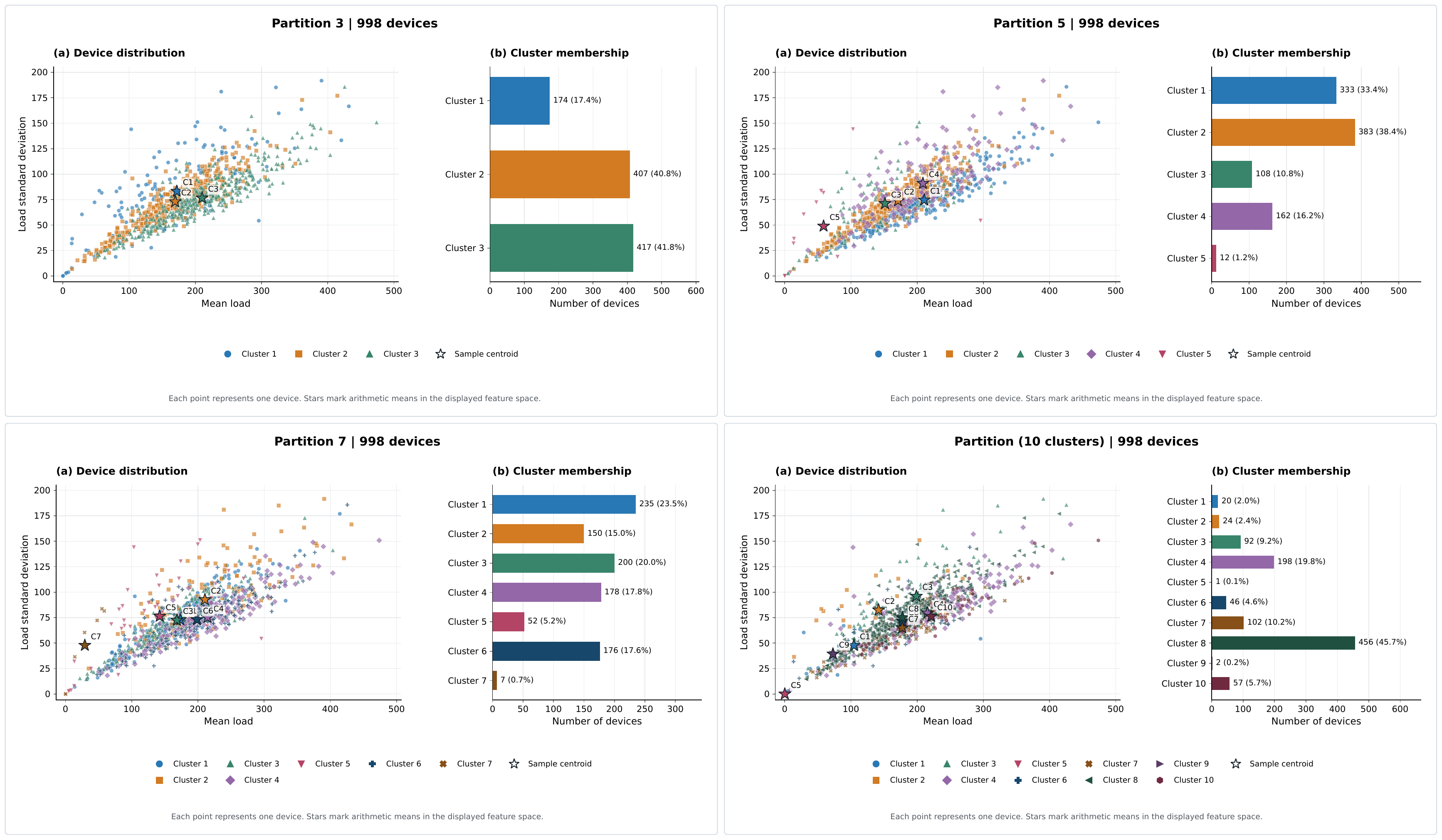}
    \caption{Clustering Result}
    \label{fig:cluster}
\end{figure}

As shown in Table~\ref{tab:gmm_cluster_sensitivity} and Figure~\ref{fig:clustering}, the validation-selected $K=7$
configuration also achieves the best test-set result for all five reported
forecasting metrics. This agreement between validation-based selection and
test-set sensitivity provides additional evidence that the selected clustering
granularity is appropriate for the present dataset. Moreover, the device
clustering results are shown in Figure~\ref{fig:cluster}. Increasing $K$ from 3 to 7 generally improves forecasting
performance, suggesting that the coarser partitions do not sufficiently
separate transformers with different temporal load characteristics. In such
cases, a cluster-specific forecasting model must still fit several distinct
load patterns simultaneously, which weakens the intended benefit of
behavior-based grouping.

The results do not continue to improve when the number of clusters is
increased from 7 to 10. Although the MSE for $K=10$ remains relatively close
to that for $K=7$, MAE and RMSE increase, MAPE rises substantially from
0.0950 to 0.2288, and MSPE increases sharply. This behavior suggests that
excessive subdivision of the transformer population can reduce the robustness
of shared forecasting models. When clusters become too small or too
specialized, the representative training subset within an individual cluster
contains less load-pattern diversity and fewer transferable training examples.

The results therefore indicate that clustering granularity involves a balance
between intra-cluster homogeneity and sufficient shared training information.
Among the evaluated settings, $K=7$ provides the most favorable balance,
suggesting that an intermediate level of model sharing is better suited to the
heterogeneous load patterns considered in this study.

\subsection{Ablation Study on Behavior-Based Clustering}
\label{subsec:clustering_ablation}

To evaluate the contribution of behavior-based transformer organization, a
random-grouping ablation is conducted. For each value of $K$, transformers
are randomly assigned to groups while preserving the same group sizes as
those produced by T-Rep--GMM clustering. The representative-device sampling,
PatchTST architecture, training procedure, and evaluation protocol are kept
unchanged.

\begin{table}[pos=htbp]
\centering
\caption{Ablation comparison between random grouping and behavior-based
clustering.}
\label{tab:clustering_ablation}
\small
\setlength{\tabcolsep}{5pt}
\renewcommand{\arraystretch}{1.12}

\begin{tabular}{llccccc}
\toprule
\textbf{Method} &
\textbf{$K$} &
\textbf{MAE} &
\textbf{MSE} &
\textbf{RMSE} &
\textbf{MAPE} &
\textbf{MSPE} \\
\midrule

Random grouping & 3
& 24.0932 & 1115.9785 & 31.4118 & 0.1636 & 0.8028 \\
ClusLoad & 3
& 21.1800 & 843.5056 & 27.5411 & 0.1866 & 0.7467 \\

\addlinespace

Random grouping & 5
& 20.9016 & 834.7437 & 27.1660 & 0.1754 & 0.6772 \\
ClusLoad & 5
& 20.6024 & 869.8537 & 27.0686 & 0.1716 & 0.6872 \\

\addlinespace

Random grouping & 7
& 22.7040 & 990.5914 & 29.5407 & 0.2006 & 2.0924 \\
ClusLoad & 7
& \textbf{18.7997} & \textbf{746.7468} & \textbf{24.6080}
& \textbf{0.0950} & \textbf{0.0125} \\

\addlinespace

Random grouping & 10
& 23.8304 & 1091.7772 & 30.9658 & 0.2456 & 6.1561 \\
ClusLoad & 10
& 19.9582 & 762.6748 & 26.1099 & 0.2288 & 4.8760 \\

\midrule

PatchTST baseline & --
& 23.4300 & 1053.8578 & 30.4793 & 0.1965 & 0.1439 \\

\bottomrule
\end{tabular}
\end{table}

The ablation results show that behavior-based grouping provides a clear
advantage at the validation-selected $K=7$ configuration. Compared with
random grouping, ClusLoad reduces MAE from 22.7040 to 18.7997, MSE from
990.5914 to 746.7468, and RMSE from 29.5407 to 24.6080, corresponding to
reductions of 17.20\%, 24.62\%, and 16.70\%, respectively. More substantial
improvements are observed for the percentage-based metrics, with MAPE
decreasing from 0.2006 to 0.0950 and MSPE from 2.0924 to 0.0125.

Compared with the direct PatchTST baseline, the same $K=7$ configuration
reduces MAE, MSE, RMSE, MAPE, and MSPE by 19.76\%, 29.14\%, 19.26\%,
51.65\%, and 91.31\%, respectively. In contrast, random grouping at $K=7$
provides only limited improvement over PatchTST on MAE, MSE, and RMSE,
while performing worse on MAPE and MSPE. These results indicate that the
benefit of ClusLoad is not simply due to dividing the transformer population
into multiple groups, but is closely related to organizing model sharing
according to learned temporal load behavior.


\subsection{High-Load Warning Performance}

In addition to continuous forecasting errors, the practical usefulness of the
forecasting results is evaluated using the warning criterion defined in
Section~4. The proposed method achieves a warning accuracy of 95.66\%,
which is higher than every comparison method in
Table~\ref{tab:forecasting_comparison}. The strongest alternative is FiLM at
88.30\%, giving the proposed framework an advantage of 7.36 percentage
points. Relative to the direct PatchTST baseline, the improvement is
7.52 percentage points.

This result is important because high-load warning depends on whether the
forecast falls on the correct side of an operational threshold, rather than
only on the average numerical distance between predicted and observed
loads. Forecasting errors around this threshold can have very different
operational consequences from errors occurring during low-load periods. A
model may therefore obtain acceptable average errors while still missing
critical high-load conditions.

The simultaneous improvement in both continuous forecasting metrics and
warning accuracy suggests that trend-based organization does not merely
smooth average predictions. Instead, separating transformers according to
their temporal behaviors helps the cluster-specific models better preserve
device-level load evolution, including periods near the warning boundary.
This property is particularly relevant for applications such as inspection
planning, load transfer, and preventive operational adjustment.

The warning results also demonstrate why the application-oriented metric
provides information complementary to conventional forecasting errors.
Different baseline models exhibit different rankings under forecasting and
warning metrics, whereas the proposed framework performs consistently well
under both perspectives. This combination is desirable for practical
distribution-system operation, where forecasting accuracy is useful only if
the resulting predictions can support timely operational decisions.


\subsection{Scalability and Practical Implications}

The main motivation of the proposed framework is to reduce the modeling
burden associated with large transformer populations without reverting to a
single global model. Under an individual-device strategy, the number of
forecasting models grows linearly with the number of transformers, resulting
in $N$ separately trained and maintained models. In the proposed framework,
the number of forecasting models is reduced to the number of clusters $K$.
For the best configuration, only seven cluster-specific PatchTST models
are required.

The reduction is also reflected in the number of transformers directly used
to train the forecasting models. At most 20 transformers are sampled from
each cluster. Therefore, when $K=7$, no more than

\begin{equation}
    N_{\mathrm{rep}} \leq 20K = 140
\end{equation}

transformers contribute directly to PatchTST training, while the resulting
cluster models are subsequently applied to the full evaluated transformer
population. The forecasting stage therefore uses a substantially smaller
training-device population than a device-wise modeling strategy.

This reduction does not imply that the entire framework is computationally
free. T-Rep representation learning and GMM fitting introduce an additional
offline preprocessing stage. However, these operations are performed to
organize the transformer population, after which forecasting models can be
trained, stored, and maintained at the cluster level. For long-term
deployment, this shifts computational effort from repeatedly maintaining a
large number of device-specific forecasting models toward a smaller number
of shared models supported by an offline representation and clustering
procedure.

The experimental results further indicate that this reduction in model
quantity does not require a corresponding sacrifice in forecasting
performance. On the contrary, the seven-cluster configuration outperforms
the direct PatchTST baseline and the other forecasting models on the major
continuous error metrics while also providing the highest warning accuracy.
The proposed method therefore achieves its intended intermediate position
between two extremes: fully personalized device-wise modeling and highly
aggregated global modeling.

From an operational perspective, the cluster structure also provides a
natural mechanism for organizing model deployment. Transformers assigned to
the same temporal-pattern group can share the same forecasting service,
which simplifies model storage, invocation, and updating. The model-assignment
mechanism described in Section~3 further allows newly introduced or
reconfigured transformers to be associated with an existing cluster according
to their learned representations rather than requiring a completely new
forecasting model. This capability is particularly relevant when the number
of monitored distribution devices continues to increase.


\subsection{Explainability Analysis}
To examine the behavior of the cluster forecasting models, we conducted a post hoc attribution analysis using temporally grouped Shapley values. Each 24-hour input window was divided into six nonoverlapping blocks of four hours. Eight training windows from the corresponding cluster were used as background references, with their starting times aligned by time of day to the explained input window. The cluster assignment and forecasting model were held fixed throughout the analysis. Enumerating all 64 block coalitions yielded grouped Shapley values for the specified background distribution at each forecast step. Positive and negative contributions respectively increase and decrease the forecast relative to the background prediction, with contributions expressed in kVA.

The analysis covered 42 forecasting windows from 14 devices across seven clusters. These devices participated in representation learning and clustering but were excluded from supervised training of their respective forecasting models. The analysis characterizes model behavior relative to the selected references and does not establish physical causality.

\subsubsection{Temporal Attribution of Individual Forecasts}
\begin{figure}[htbp]
    \centering
    \includegraphics[width=1.0\linewidth]{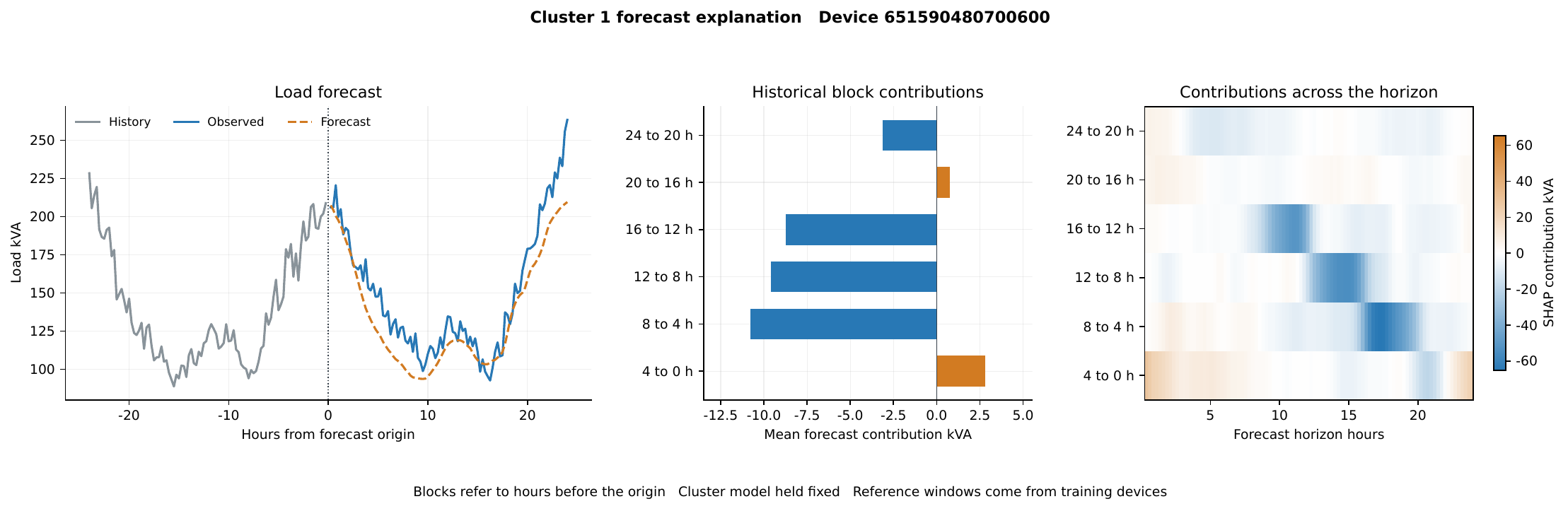}
    \caption{Example of forecast: cluster 1}
    \label{fig:cluster1_forecast}
\end{figure}
Figure 7 presents a forecasting example from Cluster 1. The model follows the broad decline and subsequent recovery in load but underestimates the rise near the end of the forecast horizon. The MAE is 6.81 kVA during forecast hours 12--16 and increases to 25.86 kVA during hours 20--24. At the final forecast step, the observed load is 263.32 kVA, compared with a prediction of 209.51 kVA. This variation demonstrates why an overall error metric should be accompanied by an examination of local deviations near peaks.

The attribution to the mean forecast indicates that several earlier historical blocks decrease the predicted load, whereas the most recent four hours provide a positive contribution. The heatmap further shows that the direction and magnitude of these contributions vary across forecast steps. At the final step, the most recent block contributes approximately +23.81 kVA, yet the forecast remains below the observation. A positive attribution therefore indicates an upward adjustment relative to the reference prediction; it does not establish that the model has captured the full magnitude of the observed increase. This example connects peak underestimation with the historical information contributing to the forecast, providing a basis for investigating the model's response to rapid load changes.

\begin{figure}[htbp]
    \centering
    \includegraphics[width=1.0\linewidth]{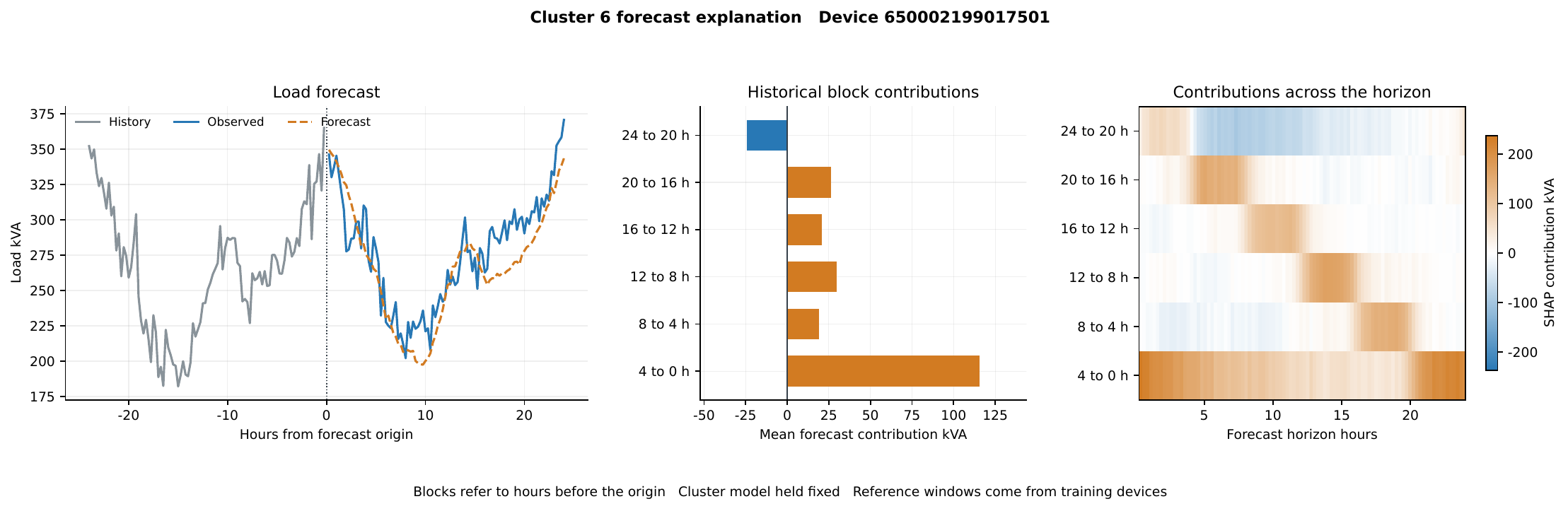}
    \caption{Example of forecast: cluster 6}
    \label{fig:cluster6_forecast}
\end{figure}

Figure 8 presents an example from Cluster 6. The model follows the overall decrease and recovery in load, although discrepancies remain in local fluctuations and the final peak. In the decomposition of the mean forecast, the most recent four hours contribute +115.80 kVA, whereas the earliest four hours contribute $-24.33$ kVA. The heatmap shows a sustained positive contribution from the most recent block across multiple forecast steps, while contributions from other historical blocks vary with the prediction horizon. For this sample, recent load information provides the largest upward adjustment, with earlier observations also contributing to the predicted trajectory.

Figures 7 and 8 illustrate different attribution patterns. However, their absolute contribution magnitudes should not be used alone to compare dependence across clusters, because device load scales and cluster background references differ.

\subsubsection{Aggregate Contributions and Explanation Diagnostics}
\begin{figure}[htbp]
    \centering
    \includegraphics[width=1.0\linewidth]{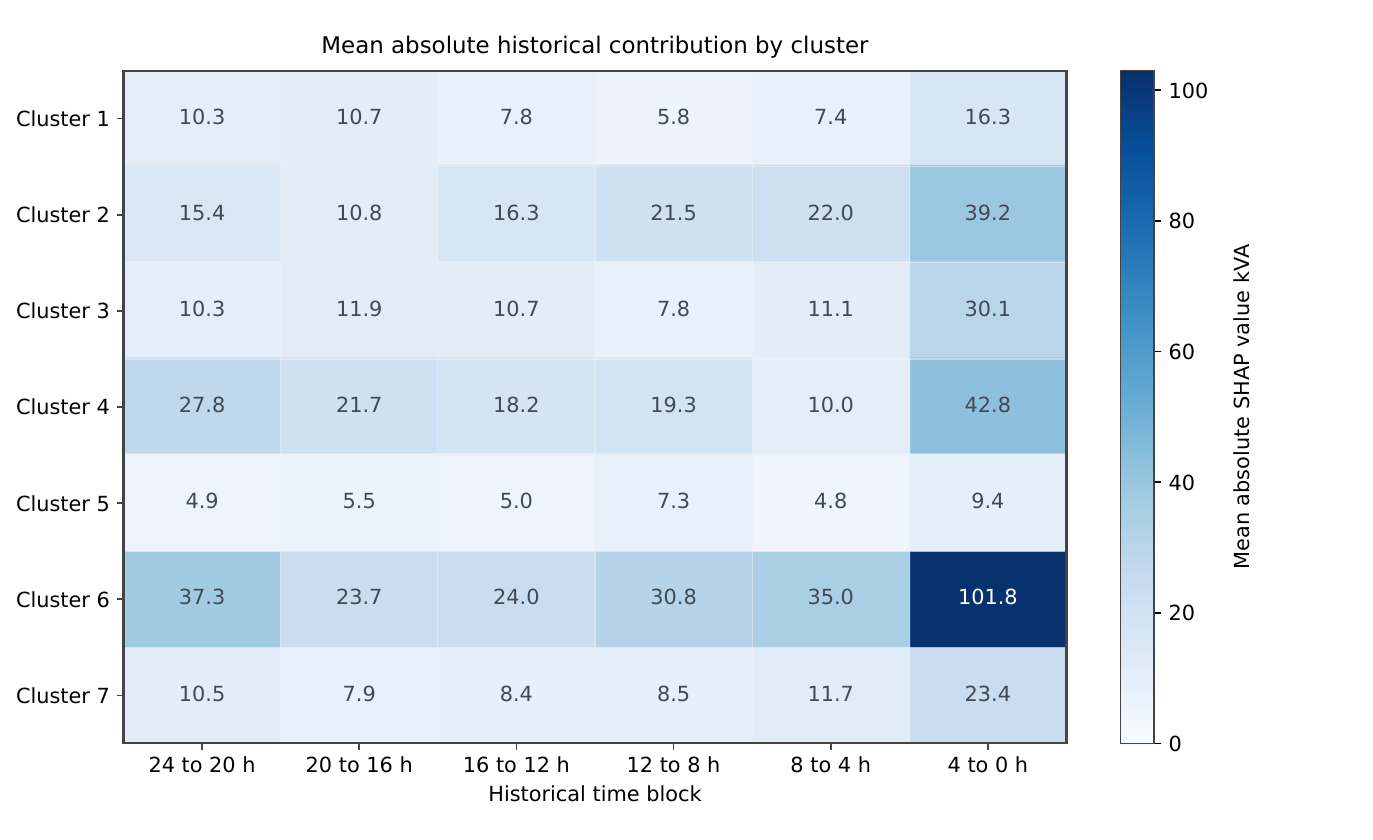}
    \caption{Mean absolute Shapley contributions}
    \label{fig:shapley_contributions}
\end{figure}
Figure 9 summarizes the mean absolute Shapley contributions across the 42 forecasting windows. Contributions are first averaged across windows within each device and then across devices within each cluster, giving each evaluated device equal weight. In the sampled cases, the most recent four-hour block has the largest mean absolute contribution in all seven clusters. Earlier blocks also contribute, indicating that predictions draw on information distributed across the historical window.

These results describe the evaluated case set. Each cluster contains only two sampled devices, so the findings should not be generalized to the entire cluster population. Larger absolute contributions also do not imply better forecasting performance or stronger physical influence.
\begin{figure}[htbp]
    \centering
    \includegraphics[width=1.0\linewidth]{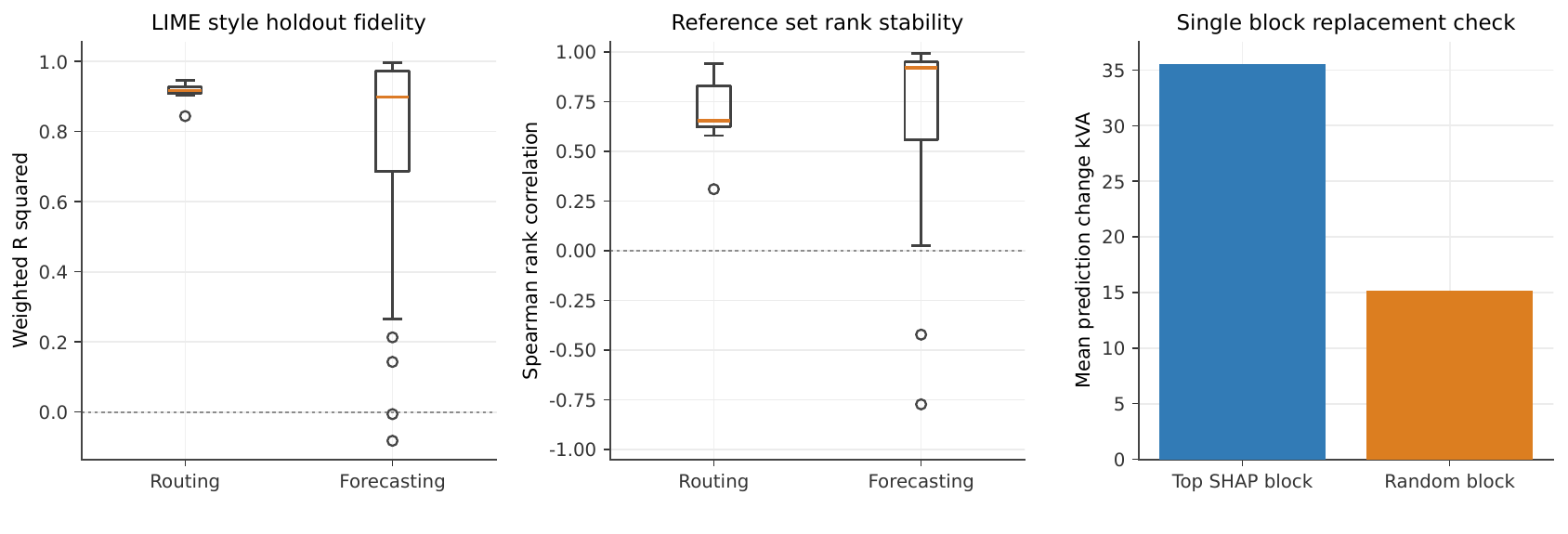}
    \caption{Local surrogate fidelity}
    \label{fig:surrogate_fidelity}
\end{figure}
Figure 10 evaluates local surrogate fidelity, sensitivity to the reference set, and prediction changes under block replacement. A local weighted linear surrogate inspired by LIME achieves a median weighted \(R^2\) of 0.898 on block combinations excluded from surrogate fitting. However, two cases have negative scores, showing that the linear approximation is inadequate for some samples under the chosen perturbation scheme.

Reducing the background from eight reference windows to four yields a median Spearman correlation of 0.914 between rankings of absolute block contributions. Although rankings are broadly consistent for many cases, two cases exhibit negative correlations, demonstrating sensitivity to reference selection. Under the same replacement scheme, replacing the block with the largest absolute contribution to the mean forecast produces a mean absolute change of 35.45 kVA in the predicted daily mean, compared with 15.22 kVA when a single block is selected uniformly at random. Highly attributed blocks therefore generally correspond to greater output sensitivity in this experiment. This diagnostic measures changes in model predictions and does not establish that replacement necessarily increases forecasting error.

\subsection{Discussion}

\subsubsection{Why Trend-Based Organization Improves Forecasting}

The observed performance gains can be understood from the way information is
shared across transformers. A global forecasting model exposes all devices to
the same parameter space regardless of their operating patterns, whereas
device-specific models eliminate information sharing altogether. The proposed
framework introduces an intermediate level of sharing: transformers with
similar historical load behaviors are grouped before forecasting, so that
model parameters are learned from devices whose temporal dynamics are more
consistent with one another.

This organization is particularly important for transformer-level forecasting,
where cross-device heterogeneity is substantial. Load curves may differ in
daily profiles, peak timing, fluctuation intensity, and short-term variability
even when transformers belong to the same geographical region. T-Rep provides
a representation space in which such temporal characteristics can be compared
more directly, while GMM separates the resulting representations into groups
with different load-pattern characteristics. The subsequent PatchTST models
therefore operate on more behaviorally coherent training populations than a
single unstructured model.

The test-set sensitivity analysis is consistent with the validation-based
selection of $K=7$ and further supports this interpretation.
The improvement from $K=3$ to $K=7$ indicates that coarse grouping still
leaves considerable heterogeneity within individual clusters. Increasing the
number of clusters beyond this point does not produce further gains, suggesting
that excessive partitioning begins to weaken the benefit of shared training
data. The role of clustering is therefore not simply to create more forecasting
models, but to identify a level of device organization at which pattern
similarity and information sharing are appropriately balanced.

\subsubsection{Interpreting the Accuracy--Scalability--Risk Relationship}

The results also reveal that forecasting accuracy, deployment scale, and
warning performance should not be treated as independent objectives. Reducing
the number of forecasting models would be of limited practical value if it
caused large losses in device-level prediction quality. Conversely, improving
average forecasting errors would provide limited operational benefit if the
resulting model failed to identify high-load conditions near the warning
threshold.

The proposed framework performs favorably under both perspectives. The
cluster-level strategy reduces the number of independently maintained
forecasting models while achieving lower MAE, MSE, RMSE, and MAPE than the
direct PatchTST baseline. At the same time, the improvement in warning
accuracy indicates that the forecasts preserve useful information around
operationally important load levels rather than merely reducing average
deviations.

The different rankings obtained by some comparison models under continuous
error metrics and warning accuracy further illustrate this distinction.
A model with competitive average errors does not necessarily provide the
most reliable threshold-event identification. This observation supports the
joint use of conventional forecasting metrics and an engineering-oriented
warning criterion when evaluating transformer-level forecasting systems.

Taken together, the results suggest that the main benefit of the proposed
framework lies in how forecasting models are organized across a heterogeneous
device population. Temporal representation learning and clustering provide
a structured mechanism for determining which transformers should share a
model, while cluster-level forecasting preserves device-level outputs. This
offers a more balanced form of model sharing than either fully global or
fully device-specific modeling.


\section{Conclusions}
\label{sec:conclusions}

This study developed ClusLoad for large-scale device-level load forecasting of distribution transformers, with the aim of balancing forecasting accuracy, cross-device heterogeneity, and model-management scalability. Rather than assigning an independent model to each transformer or using a single global model for the entire population, ClusLoad groups transformers according to their learned temporal load behaviors and allows devices with similar patterns to share forecasting models. The framework combines T-Rep-based representation learning, GMM-based clustering, representative sampling, and cluster-specific PatchTST forecasting as a device-organization and model-sharing strategy.

Experiments on real-world data from nearly one thousand distribution transformers demonstrate the effectiveness of this approach. The $K=7$ configuration selected by validation MAE also achieves the strongest test-set performance among the evaluated clustering granularities, with an MAE of 18.7997 and an RMSE of 24.6080. Compared with the direct PatchTST baseline, it reduces MAE, RMSE, and MAPE by 19.76\%, 19.26\%, and 51.65\%, respectively. The rated-capacity-based warning evaluation further shows a warning accuracy of 95.66\%, compared with 88.14\% for the direct PatchTST baseline, indicating that the improvement in numerical forecasting also benefits the identification of high-load operating states.

ClusLoad also reduces the model-management burden associated with large transformer populations. Under the best configuration, only seven cluster-specific forecasting models are maintained, while no more than 140 representative transformers are directly involved in supervised forecasting-model training. These models can then be applied to the full evaluated population while preserving device-level predictions. This intermediate modeling granularity provides a practical alternative to both fully device-specific and fully global forecasting, particularly for distribution systems where large numbers of heterogeneous transformers must be monitored and updated simultaneously.

The results also show that clustering granularity affects the balance between device specialization and model sharing, and the best configuration may vary across transformer populations and operating conditions. Future work will therefore investigate more informative representative-selection strategies, adaptive model assignment under changing load patterns, and validation across independent distribution networks. Deployment-oriented evaluations of training time, inference cost, memory use, and model-update requirements would further clarify the practical scalability of cluster-level forecasting. Overall, ClusLoad provides a practical framework for organizing heterogeneous transformer populations while jointly considering forecasting performance, deployment scalability, and high-load operational awareness.

\section*{Data and Code Availability}
The transformer operational data used in this study are associated with a confidential project and cannot be made publicly available under the data-use agreement. To support methodological reproducibility, the authors will release a refactored version of the experimental code after publication of the paper. The public release will exclude data and sensitive information subject to the confidentiality agreement.



\begin{thebibliography}{31}
\providecommand{\natexlab}[1]{#1}
\providecommand{\url}[1]{\texttt{#1}}
\expandafter\ifx\csname urlstyle\endcsname\relax
  \providecommand{\doi}[1]{doi: #1}\else
  \providecommand{\doi}{doi: \begingroup \urlstyle{rm}\Url}\fi

\bibitem[Pinheiro et~al.(2023)Pinheiro, Madeira, and Francisco]{pinheiro2023secondary}
M.~G. Pinheiro, S.~C. Madeira, and A.~P. Francisco.
\newblock Short-term electricity load forecasting---a systematic approach from system level to secondary substations.
\newblock \emph{Applied Energy}, 332:\penalty0 120493, 2023.
\newblock \doi{10.1016/j.apenergy.2022.120493}.

\bibitem[Visakh and Selvan(2023)]{Visakh2023}
A.~Visakh and M.~P. Selvan.
\newblock Analysis and mitigation of the impact of electric vehicle charging on service disruption of distribution transformers.
\newblock \emph{Sustainable Energy, Grids and Networks}, 35:\penalty0 101096, 2023.
\newblock \doi{10.1016/j.segan.2023.101096}.

\bibitem[{IEEE}(2026)]{IEEE_C5791_2025}
{IEEE}.
\newblock \emph{{IEEE} Guide for Loading Mineral-Oil-Immersed Transformers and Step-Voltage Regulators}.
\newblock IEEE, 2026.
\newblock URL \url{https://standards.ieee.org/ieee/C57.91/7163/}.
\newblock IEEE Std C57.91-2025.

\bibitem[Montero-Manso and Hyndman(2021)]{MonteroManso2021}
P.~Montero-Manso and R.~J. Hyndman.
\newblock Principles and algorithms for forecasting groups of time series: Locality and globality.
\newblock \emph{International Journal of Forecasting}, 37\penalty0 (4):\penalty0 1632--1653, 2021.
\newblock \doi{10.1016/j.ijforecast.2021.03.004}.

\bibitem[Godahewa et~al.(2021)Godahewa, Bandara, Webb, Smyl, and Bergmeir]{Godahewa2021}
R.~Godahewa, K.~Bandara, G.~I. Webb, S.~Smyl, and C.~Bergmeir.
\newblock Ensembles of localised models for time series forecasting.
\newblock \emph{Knowledge-Based Systems}, 233:\penalty0 107518, 2021.
\newblock \doi{10.1016/j.knosys.2021.107518}.

\bibitem[Grabner et~al.(2023)Grabner, Wang, Wen, Blazic, and Struc]{Grabner2023}
M.~Grabner, Y.~Wang, Q.~Wen, B.~Blazic, and V.~Struc.
\newblock A global modeling framework for load forecasting in distribution networks.
\newblock \emph{IEEE Transactions on Smart Grid}, 14\penalty0 (6):\penalty0 4927--4941, 2023.
\newblock \doi{10.1109/TSG.2023.3264525}.

\bibitem[Li et~al.(2021{\natexlab{a}})Li, Meinrenken, Modi, and Culligan]{li2021apartment}
L.~Li, C.~J. Meinrenken, V.~Modi, and P.~J. Culligan.
\newblock Short-term apartment-level load forecasting using a modified neural network with selected auto-regressive features.
\newblock \emph{Applied Energy}, 287:\penalty0 116509, 2021{\natexlab{a}}.
\newblock \doi{10.1016/j.apenergy.2021.116509}.

\bibitem[Motlagh et~al.(2019)Motlagh, Berry, and O'Neil]{motlagh2019clustering}
O.~Motlagh, A.~Berry, and L.~O'Neil.
\newblock Clustering of residential electricity customers using load time series.
\newblock \emph{Applied Energy}, 237:\penalty0 11--24, 2019.
\newblock \doi{10.1016/j.apenergy.2018.12.063}.

\bibitem[Michalakopoulos et~al.(2024)Michalakopoulos, Sarmas, Papias, Skaloumpakas, Marinakis, and Doukas]{michalakopoulos2024clustering}
V.~Michalakopoulos, E.~Sarmas, I.~Papias, P.~Skaloumpakas, V.~Marinakis, and H.~Doukas.
\newblock A machine learning-based framework for clustering residential electricity load profiles to enhance demand response programs.
\newblock \emph{Applied Energy}, 361:\penalty0 122943, 2024.
\newblock \doi{10.1016/j.apenergy.2024.122943}.

\bibitem[Yue et~al.(2022)Yue, Wang, Duan, Yang, Huang, Tong, and Xu]{yue2022ts2vec}
Z.~Yue, Y.~Wang, J.~Duan, T.~Yang, C.~Huang, Y.~Tong, and B.~Xu.
\newblock {TS2Vec}: Towards universal representation of time series.
\newblock \emph{Proceedings of the AAAI Conference on Artificial Intelligence}, 36\penalty0 (8):\penalty0 8980--8987, 2022.
\newblock \doi{10.1609/aaai.v36i8.20881}.

\bibitem[Fraikin et~al.(2024)Fraikin, Bennetot, and Allassonni{\`e}re]{fraikin2024trep}
A.~F. Fraikin, A.~Bennetot, and S.~Allassonni{\`e}re.
\newblock {T-Rep}: Representation learning for time series using time-embeddings.
\newblock In \emph{International Conference on Learning Representations}, 2024.
\newblock \doi{10.48550/arXiv.2310.04486}.

\bibitem[Fan et~al.(2024)Fan, Chen, Mo, and Liao]{fan2024pfl}
C.~Fan, R.~Chen, J.~Mo, and L.~Liao.
\newblock Personalized federated learning for cross-building energy knowledge sharing: Cost-effective strategies and model architectures.
\newblock \emph{Applied Energy}, 362:\penalty0 123016, 2024.
\newblock \doi{10.1016/j.apenergy.2024.123016}.

\bibitem[Choi et~al.(2026)Choi, Lee, Langtry, and Choudhary]{choi2026pptl}
W.~Choi, S.~Lee, M.~Langtry, and R.~Choudhary.
\newblock Privacy-preserving transfer learning framework for building energy forecasting with fully anonymized data.
\newblock \emph{Applied Energy}, 412:\penalty0 127600, 2026.
\newblock \doi{10.1016/j.apenergy.2026.127600}.

\bibitem[Ramos et~al.(2026)Ramos, Silva, Band{\'o}ria, Dias, Villela, and Morais]{ramos2026timevae}
P.~V.~B. Ramos, W.~N. Silva, L.~H.~T. Band{\'o}ria, B.~H. Dias, S.~M. Villela, and H.~Morais.
\newblock Federated electricity demand forecasting: A client selection mechanism based on {TimeVAE} latent representations.
\newblock \emph{Applied Energy}, 419:\penalty0 128064, 2026.
\newblock \doi{10.1016/j.apenergy.2026.128064}.

\bibitem[Huyghues-Beaufond et~al.(2020)Huyghues-Beaufond, Tindemans, Falugi, Sun, and Strbac]{huyghues2020cleansing}
N.~Huyghues-Beaufond, S.~Tindemans, P.~Falugi, M.~Sun, and G.~Strbac.
\newblock Robust and automatic data cleansing method for short-term load forecasting of distribution feeders.
\newblock \emph{Applied Energy}, 261:\penalty0 114405, 2020.
\newblock \doi{10.1016/j.apenergy.2019.114405}.

\bibitem[Chitalia et~al.(2020)Chitalia, Pipattanasomporn, Garg, and Rahman]{chitalia2020rnn}
G.~Chitalia, M.~Pipattanasomporn, V.~Garg, and S.~Rahman.
\newblock Robust short-term electrical load forecasting framework for commercial buildings using deep recurrent neural networks.
\newblock \emph{Applied Energy}, 278:\penalty0 115410, 2020.
\newblock \doi{10.1016/j.apenergy.2020.115410}.

\bibitem[Gao et~al.(2023)Gao, Chen, Hu, and Zhang]{gao2023adaptive}
J.~Gao, Y.~Chen, W.~Hu, and D.~Zhang.
\newblock An adaptive deep-learning load forecasting framework by integrating {Transformer} and domain knowledge.
\newblock \emph{Advances in Applied Energy}, 10:\penalty0 100142, 2023.
\newblock \doi{10.1016/j.adapen.2023.100142}.

\bibitem[Haben et~al.(2021)Haben, Arora, Giasemidis, Voss, and Vukadinovi{\'c}~Greetham]{haben2021review}
S.~Haben, S.~Arora, G.~Giasemidis, M.~Voss, and D.~Vukadinovi{\'c}~Greetham.
\newblock Review of low voltage load forecasting: Methods, applications, and recommendations.
\newblock \emph{Applied Energy}, 304:\penalty0 117798, 2021.
\newblock \doi{10.1016/j.apenergy.2021.117798}.

\bibitem[Li et~al.(2021{\natexlab{b}})Li, Zhang, Hu, and Lu]{li2021meta}
Y.~Li, S.~Zhang, R.~Hu, and N.~Lu.
\newblock A meta-learning based distribution system load forecasting model selection framework.
\newblock \emph{Applied Energy}, 294:\penalty0 116991, 2021{\natexlab{b}}.
\newblock \doi{10.1016/j.apenergy.2021.116991}.

\bibitem[Zhou et~al.(2021)Zhou, Zhang, Peng, Zhang, Li, Xiong, and Zhang]{zhou2021informer}
H.~Zhou, S.~Zhang, J.~Peng, S.~Zhang, J.~Li, H.~Xiong, and W.~Zhang.
\newblock {Informer}: Beyond efficient {Transformer} for long sequence time-series forecasting.
\newblock \emph{Proceedings of the AAAI Conference on Artificial Intelligence}, 35\penalty0 (12):\penalty0 11106--11115, 2021.
\newblock \doi{10.1609/aaai.v35i12.17325}.

\bibitem[Liu et~al.(2022)Liu, Yu, Liao, Li, Lin, Liu, and Dustdar]{liu2022pyraformer}
S.~Liu, H.~Yu, C.~Liao, J.~Li, W.~Lin, A.~X. Liu, and S.~Dustdar.
\newblock {Pyraformer}: Low-complexity pyramidal attention for long-range time series modeling and forecasting.
\newblock In \emph{International Conference on Learning Representations}, 2022.
\newblock URL \url{https://dsg.tuwien.ac.at/team/sd/papers/ICLR_2022_SD_Pyraformer.pdf}.

\bibitem[Zhou et~al.(2022{\natexlab{a}})Zhou, Ma, Wen, Wang, Sun, and Jin]{zhou2022fedformer}
T.~Zhou, Z.~Ma, Q.~Wen, X.~Wang, L.~Sun, and R.~Jin.
\newblock {FEDformer}: Frequency enhanced decomposed {Transformer} for long-term series forecasting.
\newblock In \emph{Proceedings of the 39th International Conference on Machine Learning}, volume 162 of \emph{Proceedings of Machine Learning Research}, pages 27268--27286, 2022{\natexlab{a}}.
\newblock URL \url{https://proceedings.mlr.press/v162/zhou22g/zhou22g.pdf}.

\bibitem[Zhou et~al.(2022{\natexlab{b}})Zhou, Ma, Wang, Wen, Sun, Yao, Yin, and Jin]{zhou2022film}
T.~Zhou, Z.~Ma, X.~Wang, Q.~Wen, L.~Sun, T.~Yao, W.~Yin, and R.~Jin.
\newblock {FiLM}: Frequency improved legendre memory model for long-term time series forecasting.
\newblock In \emph{Advances in Neural Information Processing Systems}, volume~35, pages 12677--12690, 2022{\natexlab{b}}.
\newblock \doi{10.52202/068431-0921}.

\bibitem[Wu et~al.(2023)Wu, Hu, Liu, Zhou, Wang, and Long]{wu2023timesnet}
H.~Wu, T.~Hu, Y.~Liu, H.~Zhou, J.~Wang, and M.~Long.
\newblock {TimesNet}: Temporal {2D}-variation modeling for general time series analysis.
\newblock In \emph{International Conference on Learning Representations}, 2023.
\newblock URL \url{https://iclr.cc/virtual/2023/poster/11976}.

\bibitem[Zeng et~al.(2023)Zeng, Chen, Zhang, and Xu]{zeng2023linear}
A.~Zeng, M.~Chen, L.~Zhang, and Q.~Xu.
\newblock Are {Transformers} effective for time series forecasting?
\newblock \emph{Proceedings of the AAAI Conference on Artificial Intelligence}, 37\penalty0 (9):\penalty0 11121--11128, 2023.
\newblock \doi{10.1609/aaai.v37i9.26317}.

\bibitem[Liu et~al.(2024)Liu, Hu, Zhang, Wu, Wang, Ma, and Long]{liu2024itransformer}
Y.~Liu, T.~Hu, H.~Zhang, H.~Wu, S.~Wang, L.~Ma, and M.~Long.
\newblock {iTransformer}: Inverted {Transformers} are effective for time series forecasting.
\newblock In \emph{International Conference on Learning Representations}, 2024.
\newblock \doi{10.48550/arXiv.2310.06625}.

\bibitem[Nie et~al.(2023)Nie, Nguyen, Sinthong, and Kalagnanam]{nie2023patchtst}
Y.~Nie, N.~H. Nguyen, P.~Sinthong, and J.~Kalagnanam.
\newblock A time series is worth 64 words: Long-term forecasting with {Transformers}.
\newblock In \emph{International Conference on Learning Representations}, 2023.
\newblock \doi{10.48550/arXiv.2211.14730}.

\bibitem[Rhodes et~al.(2014)Rhodes, Cole, Upshaw, Edgar, and Webber]{rhodes2014clustering}
J.~D. Rhodes, W.~J. Cole, C.~R. Upshaw, T.~F. Edgar, and M.~E. Webber.
\newblock Clustering analysis of residential electricity demand profiles.
\newblock \emph{Applied Energy}, 135:\penalty0 461--471, 2014.
\newblock \doi{10.1016/j.apenergy.2014.08.111}.

\bibitem[Li et~al.(2021{\natexlab{c}})Li, Gong, Fan, Peng, Chun, and Fang]{li2021crossscale}
W.~Li, G.~Gong, H.~Fan, P.~Peng, L.~Chun, and X.~Fang.
\newblock A clustering-based approach for {``cross-scale''} load prediction on building level in {HVAC} systems.
\newblock \emph{Applied Energy}, 282:\penalty0 116223, 2021{\natexlab{c}}.
\newblock \doi{10.1016/j.apenergy.2020.116223}.

\bibitem[Dong et~al.(2025)Dong, Luo, Yuan, Tian, Zhang, Wu, and Liu]{dong2025spatiotemporal}
X.~Dong, Y.~Luo, S.~Yuan, Z.~Tian, L.~Zhang, X.~Wu, and B.~Liu.
\newblock Building electricity load forecasting based on spatiotemporal correlation and electricity consumption behavior information.
\newblock \emph{Applied Energy}, 377:\penalty0 124580, 2025.
\newblock \doi{10.1016/j.apenergy.2024.124580}.

\bibitem[Fan et~al.(2020)Fan, Sun, Xiao, Ma, Lee, Wang, and Tseng]{fan2020transfer}
C.~Fan, Y.~Sun, F.~Xiao, J.~Ma, D.~Lee, J.~Wang, and Y.~C. Tseng.
\newblock Statistical investigations of transfer learning-based methodology for short-term building energy predictions.
\newblock \emph{Applied Energy}, 262:\penalty0 114499, 2020.
\newblock \doi{10.1016/j.apenergy.2020.114499}.

\end{thebibliography}
\end{document}